\documentclass[11pt]{article}

\usepackage{fullpage,amsmath, amssymb, amsfonts, amsthm, mathrsfs,mathtools,enumerate, enumitem,indentfirst}
\usepackage{bbm}
\usepackage{verbatim}
\usepackage{natbib}
\usepackage{float}
\usepackage{tikz}
\usetikzlibrary{arrows.meta}
\usepackage{xcolor}

\usepackage{braket}
\usepackage{algorithm}
\usepackage{algpseudocode}
\theoremstyle{definition}
\usepackage{hyperref}
\usepackage{cleveref}
\newtheorem{theorem}{Theorem}
\newtheorem{lemma}{Lemma}

\newtheorem{definition}{Definition}

\newtheorem{remark}[theorem]{Remark}

\crefname{assumption}{Assumption}{Assumptions}
\Crefname{assumption}{Assumption}{Assumptions}
\crefname{theorem}{Theorem}{Theorems}
\Crefname{theorem}{Theorem}{Theorems}
\crefname{lemma}{Lemma}{Lemmas}
\Crefname{lemma}{Lemma}{Lemmas}
\crefname{proposition}{Proposition}{Propositions}
\Crefname{proposition}{Proposition}{Propositions}
\crefname{example}{Example}{Examples}
\Crefname{example}{Example}{Examples}
\crefname{corollary}{Corollary}{Corollaries}
\Crefname{corollary}{Corollary}{Corollaries}
\crefname{definition}{Definition}{Definitions}
\Crefname{definition}{Definition}{Definitions}
\crefname{section}{Section}{Sections}
\Crefname{section}{Section}{Sections}
\crefname{claim}{Claim}{Claims}
\Crefname{claim}{Claim}{Claims}
\crefname{algorithm}{Algorithm}{Algorithms}
\Crefname{algorithm}{Algorithm}{Algorithms}
\crefname{appendix}{Appendix}{Appendices}
\Crefname{appendix}{Appendix}{Appendices}
\DeclareMathOperator*{\argmax}{arg\,max}

\title{Quasipolynomial-Time Planning for Decodable POMDPs}
\author{
Oliver Dobson$^{1}$ \and
Varun Jog$^{1}$ \and
Chun Hei Michael  Shiu$^{2}$
}

\date{
$^{1}$Department of Pure Mathematics and Mathematical Statistics, University of Cambridge \\
$^{2}$Department of Electrical and Computer Engineering, University of British Columbia
\vspace{0.5em}
\texttt{od290@cam.ac.uk, vj270@cam.ac.uk, shiuchm@student.ubc.ca}
}

\makeatletter
\newcommand{\skipitems}[1]{%
  \addtocounter{\@enumctr}{#1}%
}

\usepackage[left=1in, right=1in, top=1in, bottom=1in]{geometry} 

\begin{document}
\maketitle
\begin{abstract}
    We study planning for finite-horizon partially observable Markov decision processes (POMDPs) with finite state, action, and observation spaces. Both exact and approximate planning are known to be computationally hard in this setting, so prior work has explored natural assumptions that make planning easier. Recent work \citep{gol22} gave quasipolynomial-time approximate planning algorithms for \emph{observable} POMDPs. These are POMDPs whose observation channels satisfy an $\ell_1$-\emph{observability} condition, but notably, no assumptions are required on the transition dynamics. In a similar vein, for every fixed $\rho>0$, we give quasipolynomial-time approximate planning algorithms for $\rho$-\emph{decodable} POMDPs. We show that decodability is implied by both $\ell_1$- and $\ell_2$-observability and is strictly weaker than observability. Moreover, we show that the optimal decodability constant can be computed in polynomial time, whereas computing the optimal $\ell_1$-observability constant is NP-hard in the worst case. We also show that the fast filter-forgetting guarantee underlying the approach of \cite{gol22} does not extend directly from $\ell_1$-observability to constant $\ell_2$-observability. Thus, decodability is an easy-to-check, unifying assumption under which quasipolynomial-time approximate planning for POMDPs is possible.
\end{abstract}

\section{Introduction} \label{sec: introduction}

Markov decision processes (MDPs) are models used to study sequential decision making. An MDP is essentially a Markov chain whose transition matrix is determined by a user-specified action. The user receives a reward which is a function of the current state of the chain. By suitably choosing their actions depending on the current state, a user can steer the chain to reap the highest cumulative reward over some finite horizon. When the transition matrices corresponding to each action and the reward function are fully known, the \emph{planning} problem asks for the optimal rule for a user's actions. Planning for MDPs can be solved elegantly and efficiently using dynamic programming~\citep{B57}. In particular, for a discrete MDP with state space, action space, and horizon bounded by $K$, planning takes $O(K^4)$ runtime.

The \emph{partially observable} MDP (POMDP) model is obtained by a small tweak to the MDP: instead of observing the state directly, the user only observes a noisy version of it. This is a natural model for many real-world settings. Some examples include playing the game of Battleship, where each shot gives partial information about the opponent's true layout~\citep{SV10}; deciding when to feed a baby, as the hidden state ``hungry'' has to be guessed from history (time since the last feed) and mood (fussiness)~\citep{KWW22}; navigating a robot in uncertain environments~\citep{CKK96}; and treating a medical condition using a patient's medical history~\citep{HF00}.

Although POMDPs are closely related to MDPs, exact and approximate planning for POMDPs is significantly more challenging. For simplicity, consider POMDPs with discrete and finite state, action, and observation spaces, and finite horizon, all of size $K$. In both MDPs and POMDPs, the distribution of the next state depends only on the current state and action. Thus, the optimal plan for an MDP may ignore the history and use only the current state and time. This luxury is unavailable for POMDPs: the current observation does not pin down the state, and so the planner may need to consider the entire history before suggesting a new action. This is called the \emph{curse of history}, as the set of possible histories grows exponentially in the horizon $K$~\citep{PGT03}. Because the posterior distribution of the current state given the complete action-observation history is a sufficient statistic, a POMDP can be reformulated as an MDP whose state space consists of \emph{beliefs}; i.e., the posterior distribution of the current state given the entire action-observation history~\citep{KLC98}. The simplex of belief distributions may then be discretised to arrive at an approximately optimal plan~\citep{L91}. Unfortunately, this creates a new problem as a sufficiently small discretisation of the simplex has size exponential in $K$, so naïve dynamic programming has a runtime guarantee that is also exponential in $K$. This is the \emph{curse of dimensionality}~\citep{PGT03}.

Is there a way to circumvent these two curses and plan for POMDPs in time sub-exponential in $K$? Planning for POMDPs has been intensely studied since the 1960s~\citep{A65}. Much of the literature, particularly in the optimisation and operations research communities, has focused on proposing algorithms for exact or approximate planning~\citep{SS73,L91,PGT03,SS04,SV10}. However, these algorithms either have no sub-exponential runtime guarantees or no approximation guarantees. This is not too surprising in light of numerous complexity-theory results. \cite{PT87} showed that exact planning is \textsf{PSPACE}-hard even with time-homogeneous dynamics and deterministic observations. The simpler problem of finding optimal plans within reasonably restricted classes is also hard. \cite{MGLA00} showed that finding the optimal plan that uses only the current observation and time, or even just the current observation, is \textsf{NP}-hard. Similar results hold for approximate planning, for both relative error~\cite{LGM01} and additive error~\cite{gol22}.

Although these complexity-theoretic results are discouraging, they do not imply a complete dead end for the research question of efficient planning for POMDPs. After all, many problems that are \textsf{NP}-hard in the worst case can be efficiently solved in practice, as real-world problems often have more rigid structural constraints. For POMDP planning, this motivates the search for natural and well-motivated conditions under which efficient planning is possible. Prior work in this area falls in two broad categories: assume the belief space has small complexity (circumventing the curse of dimensionality), or assume the distant history is irrelevant (circumventing the curse of history). A paper in the first category is  \cite{bur96}, which assumes deterministic observations such that the preimage of each observation is a small set. This restricts the belief distribution to a smaller support and allows for an efficient discretisation. \cite{HLR07} reach a similar conclusion by assuming that the set of all possible reachable beliefs has a small cover at the required resolution. Research in the second category relies on \emph{filter stability}. A \emph{filter} is simply the rule for updating the current belief after taking an action and seeing a new observation, using the transition model and Bayes' rule. Informally, a filter is stable if it forgets its initialisation: starting from different initial beliefs and repeatedly applying the filter produces beliefs that become close over time. If this convergence is sufficiently fast, the current belief can be accurately approximated using only a short recent history, since events further back in history have little influence. Examples of works along these lines are \cite{BK98,kar22,mcd23,gol22}.

Our work in this paper is motivated by the \emph{observability} assumption in \cite{gol22} (see also \cite{EKM07}), which implies filter stability and  enables efficient (quasi-polynomial time) approximate planning for POMDPs. Observability is defined as follows:
 
\begin{definition}[$\ell_1$-observability, \citep{gol22}] \label{definition: l1-observability}
        Let $\gamma > 0$. The observation channel $K \in \mathbb{R}^{O \times S}$ from the state space $\mathcal S$ to the output space $\mathcal O$ is the matrix with entries $(K)_{y,s} = K(y|s)$. This channel is said to be $\gamma$-$\ell_1$-observable
if, for any state distributions $p, q \in \Delta(\mathcal S)$,
    \begin{align}
        \|K p - K q \|_1 \geq \gamma \|p - q\|_1. \label{eq: l1-observability}
    \end{align} 
     A POMDP satisfies $\gamma$-$\ell_1$-observability if all its observation kernels satisfy \eqref{eq: l1-observability}.
\end{definition}

Observability is therefore a reverse data processing condition on the channel, insisting that it preserve the $\ell_1$-distance between $p$ and $q$ up to a multiplicative factor of $\gamma$. The requirement that this should hold \emph{for every} $p$ and $q$ makes it stronger than it seems at first sight.

It is sensible and appealing to make assumptions on the observation process while placing no restrictions on the transition dynamics of the POMDP. As noted in \cite{EKM07,gol22}, transition dynamics may be an immutable property of a system (for example, the dynamics of a baby's hunger), but the user may be able to make their observations more informative about the hidden state (for example, by accurately measuring a baby's milk intake).

The $\ell_1$-observability assumption has two notable shortcomings. First, as noted in~\cite{gol22}, it can only hold in the \emph{undercomplete} setting; i.e., one where the number of observations is at least as large as the number of states.  And second, it is not clear if the optimal $\gamma$ can be determined efficiently for any given channel $K$. To circumvent the second shortcoming, consider the following $\alpha$-$\ell_2$-observability assumption:

\begin{definition}[$\ell_2$-observability] \label{definition: l2-observability}
    Let $\alpha > 0$. Let $K$ be as in Definition~\ref{definition: l1-observability}. We say that $K$ satisfies $\alpha$-$\ell_2$-\textit{observability} if for any distributions $p, q \in \Delta(\mathcal S)$,
    \begin{align}
        \|K p - K q \|_2 \geq \alpha \|p - q\|_2. \label{eq: l2-observability}
    \end{align}
    A POMDP satisfies $\alpha$-$\ell_2$-observability if all its observation kernels satisfy \eqref{eq: l2-observability}.
\end{definition}

Given $K$, the optimal $\alpha$ is given by $\alpha = \inf_{x \in \mathbf 1^\perp, x \neq 0} \frac{\|Kx\|_2}{\|x\|_2}$. This is the smallest singular value of $K$ restricted to the zero-sum subspace $\mathbf{1}^{\perp}$, and can be computed efficiently (polynomial time in $S$ and $O$).\footnote{The optimal $\alpha$ is at most 1. This does not follow from a data processing inequality, however, as the $\ell_2$-norm need not always contract. It follows by choosing $p = e_{s}$ and $q = e_t$ for two distinct states $s$ and $t$, and noting that the maximum $\ell_2$-distance between two probability mass functions is $\sqrt 2$.}

Could one prove exponential filter stability and therefore the existence of quasipolynomial-time approximate planning for POMDPs à la \cite{gol22} when $\ell_1$-observability is replaced by $\ell_2$-observability for constant $\alpha > 0$? \footnote{This question emerged through brief conversations the second author had with Ankur Moitra~\citep{M24}, who drew our attention to the information-theoretic approach in~\cite{gol22}, and with Noah Golowich and Dhruv Rohatgi~\citep{GR25}, who raised this specific question.} A simple norm inequality shows that an $\alpha$-$\ell_2$-observable channel is $(\alpha/\sqrt S)$-$\ell_1$ observable, but such conversion worsens the runtime guarantee of \cite{gol22} to exponential in $S$. Our work in this paper began as an effort to solve this question by establishing an $\ell_2$-observability analogue of the information-theoretic result  \cite[Lemma~4.2]{gol22}, which is the linchpin of their KL-divergence based argument. This effort ended in a productive failure and led to the present manuscript. In what follows, we highlight our main results.

\subsection{Main contributions}

\paragraph{\textbf{(1) Quasipolynomial-time planning for decodable POMDPs.}} Our principal contribution is the following definition.

\begin{definition}[$\rho$-decodability] \label{def: decodability}
    Let $\rho \in (0,1]$. Let $K$ be as in Definition~\ref{definition: l1-observability}. We say that $K$ is $\rho$-decodable if there exists a decoder channel $D$ from the observation space $\mathcal O$ to the state space $\mathcal S$ such that the cascaded channel $DK$ from $\mathcal S$ to $\mathcal S$ satisfies $(DK)_{(s, s)} \geq \rho$ for every $s \in \mathcal S$. Equivalently, $\sum_{y \in \mathcal O} K(y|s) D(s|y) \geq \rho$ for all $s \in \mathcal S$. A POMDP is said to be $\rho$-decodable if all its observation channels are $\rho$-decodable.
\end{definition}

We prove the following result:

\begin{theorem}\label{thm: main theorem}
    Let $\mathcal P$ be a $\rho$-decodable POMDP and let $\varepsilon > 0$. There is an algorithm that produces a deterministic policy whose expected cumulative reward is within $\varepsilon$ of the optimal value, and which runs in time at most $\mathrm{poly}(\text{input size})\left(\frac{C_1AOSH}{\varepsilon}\right)^{\frac{C_2\log(C_3H/\varepsilon)}{\rho}}$ for some universal constants $C_1,C_2,C_3>0$.
\end{theorem}

We give some intuition as to why this theorem holds. Suppose we expand the observation space to include both the observation and the decoder's output. If the planning algorithm can decode the state exactly at a certain time, it can simply disregard the history. But the success probability of the decoder $D$ is  $\rho < 1$. Intuitively, over a history of length $L \approx \log(1/\delta)/\rho$, there is at least one time instant where the state was correctly decoded, or \emph{revealed}, with probability at least $1-\delta$. The problem is we don't know when the revelation happened. For ease of explanation, assume revelations always occur within the past $L$ steps. If we knew the time of the most recent revelation, we could compute the current belief exactly by keeping track of the actions and observations since the revelation. Since we don't know when the revelation happened, we keep track of its probability distribution over $\{0, 1, \dots, L-1\}$, which lies in the simplex of probability vectors with $L$ coordinates. There are at most $(AOS)^L$ action-observation histories (recall that the expanded observation space is of size $OS$). Furthermore, the resolution at which the simplex needs to be approximated turns out to be $\varepsilon/H^2$ to control the cumulative error over a horizon $H$. This needs $(H/\varepsilon)^{O(L)}$ discretisation points on the simplex. Now a planning algorithm can run dynamic programming on the quantised belief state that takes at most $(AOSH/\varepsilon)^{O(L)}$ values and arrive at an approximately optimal policy. Substituting a suitable value of $\delta$ gives the final expression. This is basically the flow of the entire proof. 

It is important to note that filter stability or fast forgetting is not invoked anywhere, and indeed, the current belief may depend on the entire history. The crucial part is that it can be approximately represented using a compact summary statistic that is easy to update.

\paragraph{\textbf{(2) Observability to decodability.}}

In \cref{sec: decodability from observability} we show that decodability subsumes existing observation-channel assumptions from \cite{gol22} and \cite{bur96}, as well as the new $\ell_2$-observability assumption discussed earlier. As a consequence, \Cref{thm: main theorem} implies quasipolynomial-time approximate planning under $\ell_1$-observability, recovering the main result from~\cite{gol22} in a different way, and also extends this conclusion to $\ell_2$-observability.

To give some intuition, consider the assumption in~\cite{bur96} first. Here, the observation channel $K$ is assumed to be deterministic such that each output symbol $y \in \mathcal O$ has a preimage $\mathcal S_y$ of size at most a constant $m$. Consider the following decoder $D$: map input $y$ uniformly to $\mathcal S_y$. Clearly, this decoder is $(1/m)$-decodable, and we immediately conclude quasipolynomial-time approximate planning via \Cref{thm: main theorem}. In fact, we don't even need $K$ to be deterministic, we just need that each $y$ can result from at most $m$ inputs to $K$. The proofs showing that $\ell_1$- and $\ell_2$-observability also imply decodability are more technical, but the idea is the same: construct an explicit decoder $D$ whose success probability is at least $c_1\gamma^2$ under $\gamma$-$\ell_1$-observability, and at least $c_2\alpha^2$ under $\alpha$-$\ell_2$-observability, for universal constants
$c_1,c_2>0$.

\cite{gol22} also established a quasipolynomial lower bound
on approximate planning for $\ell_1$-observable POMDPs under the Exponential Time Hypothesis (ETH). More specifically,
\cite{gol22} construct a family of $\ell_1$-observable POMDPs with constant observability parameter for which approximate planning for $\varepsilon=0.1$ with runtime $K^{o(\log K)}$ is impossible under ETH, where
$K=\max\{H,S,A,O\}$. Since constant $\ell_1$-observability implies constant decodability, the same lower-bound
construction also applies to decodable POMDPs.

\paragraph{(3) Limitations of prior work.} We next examine the role of $\ell_1$-observability in~\cite{gol22}, identifying limitations of both the
assumption itself and the filter-stability approach to planning.

First, we give an explicit example of an $\alpha$-$\ell_2$-observable POMDP with constant $\alpha$ where the filter forgetting is too slow to directly extend the approach from~\cite{gol22}. Specifically, there exist some initial beliefs $b_0, \widehat{b}_0$ such that it requires $\Omega(S \log (1/\varepsilon))$ steps for their expected $\ell_1$-distance to fall below $\varepsilon$. The linear dependence on $S$ prevents us from obtaining the polylogarithmic memory bound needed for their quasipolynomial-time analysis (see \Cref{app: counterexample}). Second, we show that computing the optimal $\ell_1$-observability parameter of a given observation channel is NP-hard in the worst case (see \Cref{app: hardness of verifying l1-observability}). In contrast, the optimal decodability parameter and a corresponding decoder $D$ can be obtained by solving a linear programme in polynomial time. Together, these observations motivate decodability as the appropriate assumption: it is efficiently checkable and, for every fixed $\rho>0$, supports quasipolynomial-time planning without requiring rapid filter convergence.

\paragraph{Overview.}

\Cref{sec: prelim} introduces the POMDP model and its associated notation. \Cref{sec: augmented} defines and analyses a POMDP closely related to the original one, which leads to the quasipolynomial-time algorithm that is presented in \Cref{sec: quasi}. \cref{sec: decodability from observability} shows that decodability is implied by observability. We conclude the paper in \Cref{sec: conclusion}.

\paragraph{Notation.} For $n \in \mathbb{N}$, denote $[n] \triangleq \{1,2,\ldots, n\}$. For $i \leq j$, denote $[i:j] = \{i, i+1,\ldots, j\}$. For a finite set $\mathcal X$, let $\Delta(\mathcal X)$ denote the set of all probability distributions on $\mathcal X$. The $\ell_1$ and $\ell_2$ distances for $p, q \in \Delta(\mathcal X)$ are denoted by $\|p-q\|_1$ and $\|p-q\|_2$, respectively. For an event $E$, we write $\mathbf{1}\{E\}$ as the indicator of $E$. For a dimension $d$, write $\mathbf{1}_d$ as the all-one vector of dimension $d$ and $\mathbf{1}_d^\perp = \{v \in \mathbb{R}^d: \mathbf{1}_d^\top v = 0\}$ to be the zero-sum subspace. Let $e_i$ denote the standard basis vector and $\delta_i$ to be the point mass distribution at $i$. For a matrix $M$, denote its entry at row $i$, column $j$ by $M_{i,j}$ and its transpose by $M^\top$.

\section{Preliminaries} \label{sec: prelim}

A \emph{partially observable Markov decision process} (POMDP) $\mathcal{P}$ consists of a state space $\mathcal S$ of size $S$, an action space $\mathcal A$ of size $A$, an observation space $\mathcal O$ of size $O$, and a finite time horizon $H$. At time $h = 1$, the initial state $s_1$ of the POMDP is drawn from a known distribution $\mu_1$. An action $a_h \in \mathcal A$ at time $h \in [1:H-1]$ updates the state $s_h \in \mathcal S$ to $s_{h+1} \sim T_h( \cdot | s_h, a_h)$. This collection of Markov kernels is denoted by $T = \{T_h\}_{h \in [1:H-1]}$. For $h \geq 2$, the planner observes $y_h \in \mathcal O$, which is the hidden state $s_h$ passed through an observation channel $K_h$; i.e., $y_h \sim K_h( \cdot | s_h)$. The collection of observation channels is denoted by $K = \{K_h\}_{h \in [2:H]}$. Finally, the collection of reward functions $r_h: \mathcal O \to [0,1]$ is denoted by $r = \{r_h\}_{h \in [2:H]}$. The planner receives a reward $r_h(y_h)$ when the observation at time $h$ is $y_h$. The POMDP $\mathcal P$ with all its parameters is thus $\mathcal{P} = (H, \mathcal S, \mathcal A, \mathcal O, \mu_1, T, K, r)$.

The planner knows $\mu_1$, $T$, $K$, and $r$, and their goal is to design a policy  for picking actions so as to maximise the cumulative expected reward. A policy can choose the action at time $h \in [1:H-1]$ based on the entire history of actions and observations up to time $h$. This is denoted by $\mathsf{h}_h = (a_{1:h-1}, y_{2:h})$. Note that $\mathsf{h}_1 = \emptyset$. Let $\mathcal H_h$ denote the set of all histories at time $h$. A planner deploys a policy $\pi = \{\pi_h\}_{h \in [1:H-1]}$, where $\pi_h: \mathcal H_h \to \mathcal A$, to determine the action $a_h$ to be taken at time $h$. The value of policy $\pi$ in $\mathcal{P}$ is the expected cumulative reward defined as 
\begin{align*}
    V^\pi := V^\pi_{\mathcal{P}} := \sum_{h=2}^H \mathbb{E}_{\mathcal{P}, \pi} \left[ r_h(y_h)\right],
\end{align*}
where the expectation is over the initial distribution $\mu_1$, hidden state transitions, observations specified by $\mathcal{P}$, and actions sampled from $\pi$. The optimal value of a POMDP $\mathcal{P}$ is 
\begin{align*}
    V^*_{\mathcal{P}} := \sup_{\pi} V^\pi_{\mathcal{P}},
\end{align*}
where the supremum is taken over all history-dependent randomised policies. Since the state, action, observation alphabets and the horizon-length are finite, the supremum is attainable, i.e., there exists an optimal policy $\pi^*$ such that $V_{\mathcal{P}}^* = V^{\pi^*}_{\mathcal{P}}$. The exact planning problem is to find a policy that attains $V_{\mathcal{P}}^*$. An approximate planning problem with additive error $\varepsilon$, which is the focus of this work, asks for a policy satisfying $V_{\mathcal P}^\pi \geq V_{\mathcal P}^* - \varepsilon$. The class of policies considered above are \emph{deterministic}; i.e., the action at time $h$ is a function of the history $\mathsf{h}_h$. A \emph{randomised policy} may use a private random seed that it may retain and reuse throughout the execution of the policy. Fixing the seed yields a deterministic policy, and thus a randomised policy is a mixture of deterministic policies. From this, it readily follows that optimal value of the POMDP with randomised policies is the same as that with deterministic policies.

\section{Augmented and $L$-killed POMDP}\label{sec: augmented}

In this section, we start with a POMDP $\mathcal P$ that is assumed to be $\rho$-decodable. Recall that this assumption means that for each observation channel $K_h \in \mathbb{R}^{O \times S}$, there is an optimal decoder $D_h \in \mathbb{R}^{S \times O}$ such that $(D_hK_h)_{s,s} \geq \rho$ uniformly for every state $s$ and time $h$. In other words, the decoder succeeds in correctly identifying the input state with probability at least $\rho$ for each state $s$ and each time $h$. As noted in the introduction, if we knew that the decoder had correctly identified the state at any time, we could ignore the entire history. Unfortunately, we can never be certain if the decoder has correctly decoded at any time point. But we can be fairly certain that the decoder was correct at some previous time step over the most recent $L$ outputs of the decoder, for some large enough $L$. We will make this idea more precise in what follows by first defining an \emph{augmented POMDP} denoted by $\widehat{\mathcal P}$, and then an $L$-\emph{killed POMDP} denoted by $\widetilde{\mathcal P}^{(L)}$. Our goal in this section is to show that if we can approximately plan for $\widetilde{\mathcal P}^{(L)}$, then the same plan remains approximately optimal even for $\mathcal P$. This reduces the problem of quasipolynomial-time approximate planning for $\mathcal P$ to doing the same for $\widetilde{\mathcal P}^{(L)}$.

\paragraph{Augmented POMDP:} The only difference between the augmented POMDP and the original POMDP is that after we receive the observation $Y_h = y$, we pass it through the decoder and obtain $\widehat{X}_h \sim D_h(\cdot| y)$. We define the augmented observation to be $Z_h := (Y_h, \widehat{X}_h)$ with observation alphabet $\mathcal Z := \mathcal O \times \mathcal S$. To be precise, the augmented observation channel at time $h$ is 
\begin{align}
    J_h((y,\widehat{s}) | s) := K_h(y|s) D_h(\widehat{s} | y). \label{eq: augmented channel}
\end{align}
Note that the extra coordinate $\widehat{X}_h$ contains no extra information compared to $Y_h$. 

\begin{lemma}[Proved in \Cref{sec: proof of augmented value eq}]\label{lemma: augmented value equivalence}
    The augmented POMDP $\widehat{\mathcal P}$ with observation $Z_h = (Y_h, \widehat{X}_h)$ has the same optimal value as the original POMDP $\mathcal P$; i.e., ${V}^*_{\widehat{\mathcal P}} = V^*_{\mathcal P}$.
\end{lemma}

The observation channels $J_h$ admit a particularly useful decomposition into a ``reveal'' channel and a ``residual'' channel, given in the following lemma:

\begin{lemma}[Proved in \cref{sec: proof of exact-reveal decomposition}]\label{lemma: exact-reveal decomposition}
    For $2 \leq h \leq H$, there exist channels $R_h$, called the reveal channel, and $N_h$, called the residual channel, from the state alphabet $\mathcal S$ to the augmented observation alphabet $\mathcal Z := \mathcal O \times \mathcal S$ such that 
    \begin{align*}
        J_h = cR_h + (1-c) N_h \quad \text{and} \quad R_h((y,\widehat{s}) | s) > 0 \implies \widehat{s} = s.
    \end{align*}
    The mixing coefficient $c = \frac{\rho}{2}$ is uniform for all $2 \leq h \leq H$.
\end{lemma}

The reveal-residual decomposition has the following interpretation. The observation at time $h$ is generated by first sampling $F_h \sim \mathrm{Bern}(c)$ as a latent random variable, independent of all past states, actions, observations, and previous $F$. Then generate
\begin{align*}
    Z_h \sim \begin{cases}
        R_h(\cdot | X_h), & \text{if } F_h = 1, \\
        N_h(\cdot | X_h), & \text{if } F_h = 0.
    \end{cases}
\end{align*}
It is clear that $Z_h \sim J_h$ by marginalising over $F_h$. Moreover, whenever $F_h = 1$, we have $\widehat{X}_h = X_h$, and we call this \emph{exact reveal} or \emph{revelation} of the hidden state. Note that the planner does not observe $F_h$ directly. It only observes $Z_h = (Y_h, \widehat{X}_h)$, but does not know whether it is generated according to $R_h$ or $N_h$. We could modify our POMDP to have state space $\mathcal S \times \{0,1\}$ in order to make the latent Bernoulli a part of the hidden state, and suitably redefine the transition and observation matrices so that resulting POMDP is the same from the planner's perspective. We don't do this, but instead do something similar: we modify the state space to keep track of ``how many steps ago was most recent exact reveal?''

It is not hard to see that the latent variables $\{F_h\}_{h \in [2:H]}$ are i.i.d. $\mathrm{Bern}(c)$. Fix an integer $L \geq 1$, let $E_L$ be the event that among the latent variables $F_2,\ldots, F_H$, there is a run of $L$ consecutive zeros. Note that conditioned on the event $E_L^c$, we are assured that  at each time step $h$, there was an exact reveal at some (unknown) point within the prior $L$ steps. A simple union bound gives $\Pr(E_L) \leq H(1 - c)^L$. This bound is uniform over all policies and all transition kernels. By choosing $L$ to be suitably large (polylogarithmic in the parameters) we can make this error smaller than $\varepsilon/(2H)$. Consequently, an $(\varepsilon/2)$-approximately optimal plan for POMDP conditioned on $E_L$ will be $\varepsilon$-approximately optimal for $\widehat{\mathcal P}$. This is a viable proof strategy, but it makes the algebra a bit more complicated, since the distribution of $F_h$ conditioned on $E_L$ is not i.i.d.\ $\mathrm{Bern}(c)$ anymore. Instead, we introduce a killed process $\widetilde{\mathcal P}^{(L)}$ that makes the analysis a bit more simple.

\paragraph{$L$-killed POMDP:} 

The $L$-killed process $\widetilde{\mathcal P}^{(L)}$ evolves exactly as the augmented POMDP $\widehat{\mathcal P}$ until a run of $L$ consecutive residual channels are used, i.e., the event $E_L$ occurs. At the time step when $E_L$ occurs, the killed model enters an absorbing ``cemetery'' state, emits a ``cemetery'' observation, and earns zero reward from that stage onward regardless of the action. This POMDP can be defined by changing the state space to also include a new variable that keeps track of when the most recent reveal happened. The state transitions and observation channels can be adjusted in a straightforward manner so that $\widetilde{\mathcal P}^{(L)}$ evolves exactly like $\widehat{\mathcal P}$ until the time step (if ever) when $E_L$ occurs.

We now define $\widetilde{P}^{(L)}$ more precisely. Fix some integer $L \geq 1$, and let $\widehat{\mathcal P} = (H, \mathcal S, \mathcal A, \mathcal Z, \mu_1, T, J, r)$ be the augmented POMDP. The $L$-killed process is denoted by $\widetilde{\mathcal P}^{(L)} = (\widetilde{H}, \widetilde{\mathcal S} ,\widetilde{\mathcal A}, \widetilde{\mathcal O}, \widetilde{\mu}_1, \widetilde{T}, \widetilde{K}, \widetilde{r})$. The horizon-length and action alphabet of $\widetilde{\mathcal P}^{(L)}$ are the same as the original POMDP, i.e., $\widetilde{H} = H$ and $\widetilde{\mathcal A} = \mathcal A$.

While the process is alive, let $Q_h$ denote the number of steps since the most recent reveal, with $Q_h=\perp$ if no reveal has occurred yet. Thus $Q_h=0$ means that a reveal has just occurred. Whenever $Q_h\neq\perp$, survival implies
$Q_h\in\{0,\ldots,L-1\}$. Let $\mathcal Q := \{\perp, 0, 1, \ldots, L-1\}$ be the alphabet for $Q_h$, and correspondingly, define $\widetilde{\mathcal S} := (\mathcal S \times \mathcal Q) \cup \{\partial\}$ to be the state alphabet for the killed process, where $\partial$ is referred to as the cemetery or dead state. Any state that is not $\partial$; i.e., of the form $(X_h, Q_h)$, is called an \emph{alive} state.

Recall that the observation alphabet of the augmented POMDP is $\mathcal Z := \mathcal O \times \mathcal S$. The observation alphabet of the killed process is defined by including the cemetery observation $\dagger$, i.e., $\widetilde{\mathcal O} := {\mathcal Z} \cup \{\dagger\} = \left( \mathcal O \times \mathcal S \right) \cup \{\dagger\}$.

The $L$-killed POMDP works as follows. The initial hidden state is $(X_1,\perp)$, where $X_1\sim\mu_1$. Thus $\widetilde\mu_1$ is supported on
$\mathcal S\times\{\perp\}$ and agrees with $\mu_1$ on the $X$ part of the state.

Suppose the process is alive at time $h<H$, with hidden state $(X_h,Q_h)$, and action $a_h$ is taken. Generate a proposed $X_{h+1}$ according to $T_h(\cdot\mid X_h,a_h)$ and, independently, sample $F_{h+1}\sim \mathrm{Bern}(c)$. If $F_{h+1}=1$, the next state is $(X_{h+1},0)$. If $F_{h+1}=0$, the next hidden state is determined as follows. When $Q_h=r\in\{0,\ldots,L-2\}$, the new state is $(X_{h+1}, Q_h + 1)$. When
$Q_h=L-1$, sampling $F_{h+1}=0$ leads to run of $L$ zeros, which updates the state to the cemetery state $\partial$. When $Q_h=\perp$; i.e., there has been no reveal in the past $h-1$ steps, then sampling $F_{h+1} = 0$ keeps  $Q_{h+1}=\perp$ when $h<L$, and kills the process by moving to $\partial$ when $h \geq L$. Once the process enters $\partial$, it remains there regardless of future actions.

For time $h\geq2$, the observation channel depends on $Q_h$. If the hidden state is $(X_h,0)$, generate $Z_h$ from the reveal channel $R_h(\cdot\mid X_h)$. If the hidden state is $(X_h,Q_h)$ with $Q_h\in\{\perp,1,\ldots,L-1\}$, generate $Z_h$ from the residual channel $N_h(\cdot\mid X_h)$. If the hidden state is $\partial$, emit $\dagger$ with probability one.

The reward for an observation $Z_h=(Y_h,\widehat X_h)$ is the same reward
as in $\widehat{\mathcal P}$, namely the original reward associated with $Y_h$. The cemetery observation $\dagger$ receives reward zero. In particular, observing $\dagger$ tells us that the hidden state is $\partial$, and since it remains $\partial$ thereafter, the future observations and rewards are also $\dagger$ and $0$, respectively, regardless of any future actions. 

We now compare the optimal value of original POMDP $\mathcal P$ with the optimal value of the killed-POMDP $\widetilde{\mathcal P}^{(L)}$. Since the augmented POMDP agrees with the killed-POMDP until killing time, it is reasonable to expect their optimal values are close to each other for large enough $L$. This is established in the following lemma, whose proof is deferred to Appendix~\ref{appendix: killing loss}.

\begin{lemma} \label{lemma: killing loss}
    Let $V^*$ be the optimal value of the original POMDP $\mathcal P$, and equivalently, $V^*$ is the optimal value of the augmented POMDP $\widehat{\mathcal P}$. Let $V_L^*$ be the optimal value of the killed POMDP $\widetilde{\mathcal P}^{(L)}$. Then 
    \begin{align}
        V^* \leq V_L^* + H^2(1-c)^L. \label{eq: killing loss}
    \end{align}
Moreover, if $\pi_L$ is a policy for the killed POMDP, it can be extended to a policy $\pi$ for the augmented POMDP such that $V^\pi \geq V_L^{\pi_L}$. Consequently, if $V_L^{\pi_L} \geq V_L^* - \eta_0$ for some $\eta_0 \geq 0$, then $V^\pi \geq V^* - H^2(1-c)^L - \eta_0$. 
\end{lemma}

The above lemma suggests a strategy for approximate planning for ${\mathcal P}$: Choose $L$ such that $H^2(1-c)^L \leq \varepsilon/2$, and devise an $(\varepsilon/2)$-approximately optimal plan $\pi_L$  for $\widetilde{\mathcal P}^{(L)}$. Then construct the corresponding plan $\pi$ for  $\widehat{\mathcal P}$, which is assured to be $\varepsilon$-approximately optimal for $\widehat{\mathcal P}$. This plan is a randomised plan for $\mathcal P$, so a final derandomisation step (that does not decrease the value) would lead to the  desired deterministic plan for $\mathcal P$. All this needs to be done in quasipolynomial-time. The most crucial step in this approach is constructing an approximately-optimal plan $\pi_L$ for the $L$-killed POMDP in quasipolynomial-time. This is what we tackle next.

\section{Quasipolynomial-time planning algorithm}\label{sec: quasi}

We'll initially work entirely within the killed-POMDP. We'll first show how to represent the belief compactly, and propose and analyse the finite-state planner based on this representation. With suitable parameter choices, this same planner gives a (randomised) quasipolynomial-time planner for the original POMDP. A final de-randomisation step can be done without worsening the error and runtime guarantees.

\subsection{Representation of beliefs}

We first show that posterior distribution of the hidden state has a compact representation. Write $\mathsf H_h$ for the random history at time $h$ and $\mathsf h_h$ for a realisation. Let $\mathsf{h}_h$ be an alive history realisation (alive meaning its latest observation is not a $\dagger$)  with $\Pr(\mathsf{H}_h = \mathsf{h}_h) > 0$. Consider the probability distribution of the most recent revelation time given the history, given by $w_{h,q} = w_{h,q}(\mathsf{h}_h) := \Pr(Q_h = q | \mathsf{H}_h = \mathsf{h}_h).$ If $w_{h,q} > 0$, define the belief of the current state conditioned on $Q_h = q$ and $\mathsf{H_h} = \mathsf{h}_h$ as 
\begin{align}
    \phi_{h,q}(s) = \phi_{h,q}(s; \mathsf{h}_h) := \Pr(X_h = s| Q_h = q, \mathsf{H}_h = \mathsf{h}_h). \label{eq: state conditional belief}
\end{align}
If $w_{h,q}(\mathsf{h}_h)=0$, its conditional state belief does not affect the true joint belief. For use with other weight vectors (for example, a quantised version of $w_h$) we choose these distributions by the suffix-based convention described later. The joint belief of $(X_h, Q_h) = (s,q)$ given the history is 
\begin{align*}
    \beta_h(s,q) = \beta_h(s,q;\mathsf{h}_h) = w_{h,q} \cdot \phi_{h,q}(s) = \Pr(X_h = s, Q_h = q| \mathsf{H}_h = \mathsf{h}_h).
\end{align*}
There are two notable points about this formula. First, the current belief is a mixture of at most $L$ beliefs, with $\phi_{h,q}$ weighted according to $w_{h,q}$. Second, conditioned on $Q_h = q$, we know the state at time $h-q$ exactly and indeed, this information is contained in $\mathsf{h}_h$. (If $h \leq L$ and $Q_h = \perp$, then we know the starting distribution $\mu_1$ exactly.) The history also contains all subsequent actions and observations after this revelation. For each reveal age $q$ with positive posterior probability, the conditional state distribution $\phi_{h,q}$ is determined by $h$, the recent suffix, and $q$. The full joint belief $\beta_h(s,q)$ is reconstructed from these conditional distributions together with the maintained weight vector $w_h$, which accounts for the contribution of the older history. Motivated by this observation, we therefore define the length-$L$ suffix of a history as follows:

\begin{definition}[Suffix] \label{def: suffix}
    Let $\mathsf{h}_h = (A_{1:h-1}, \widetilde{Z}_{2:h})$ be the history at time $h$. Define the \textit{recent-history suffix}, or simply \textit{suffix} $u_h$ to be the entire history if $h \leq L$, and $(\widetilde{Z}_{h-L+1:h}, A_{h-L+1: h-1})$ otherwise. Thus, $u_h$ contains at most $L$ actions and at most $L$ observations. Denote the collection of all suffixes at time $h$ by $\mathcal U_h$.
\end{definition}
The following lemma, proved in~\Cref{appendix: locality} makes our observation precise:
\begin{lemma} \label{lemma: locality}
 For every alive history $\mathsf{h}_h$ and every $q \in \mathcal Q$, the conditional state belief $\phi_{h,q}$ is determined by the time $h$, the recent suffix $u_h$, and the time since last reveal $q$. Notably, it is independent of the part of the history before $u_h$.
\end{lemma}

Being able to calculate the exact belief using $(h, u_h, w_h)$ is not enough for planning. We also need to be able to update this triplet efficiently after taking action $a_h$ and observing an alive $z_{h+1}$. Updating $h$ and $u_h$ is straightforward: $h$ becomes $h+1$, and $u_h$ drops the earliest action/observation pair and replaces it with the most recent one. Our next lemma, proved in \Cref{appendix: closure of actual belief} shows that even the weight vector $w_{h+1}$ can be computed efficiently from $(h,u_h,w_h)$, $a_h$, and $z_{h+1}$ alone, and that the resulting triple represents the exact updated belief. 

A restriction is needed before applying this update to approximate weights, which is what we shall do later. For the true posterior, an arbitrary $\phi_{h,q}$ when $w_{h,q}=0$ contributes nothing. However, approximate representation (after rounding or quantising) may give $q$ positive weight. Updating this arbitrary $\phi_{h,q}$ need not give the conditional distribution assigned to the new suffix, so changing the weights alone would not represent the exact posterior. We therefore allow positive weight \emph{only} at ages for which the suffix-based filtering calculation is well-defined.

To that end, let $A_h(u)$ be the set of \emph{admissible ages}. A numerical age $q\in\{0,\ldots,L-1\}$ is admissible if $\tau=h-q\geq2$, the recorded reveal observation has $R_\tau(z_\tau\mid\widehat x_\tau)>0$, and filtering from $\delta_{\widehat x_\tau}$ through the recorded transitions and residual observations at times $\tau+1,\ldots,h$ has positive likelihood. The age $\perp$ is admissible if $h\leq L$ and filtering from $\mu_1$ through the full recorded history using residual channels has positive likelihood (including the empty history at $h=1$). For admissible ages, define $\phi_{h,q}^u$ by this calculation; otherwise set it to $\mu_1$. We always ensure that any rounding or quantising of $w_{h}$ remains supported on $A_h(u)$. Note that $A_h(u)$ has size at most $L$ and it can be efficiently computed using the prefix $u$.

\begin{lemma} \label{lemma: closure of actual belief}
Let $\beta(s,q)=\Phi_h(u,w)(s,q)=w_q\phi_{h,q}^u(s)$, where $w$ is supported on $A_h(u)$. After taking action $a$ and observing an alive observation $\widetilde z$ with positive probability under $\beta$, the exact posterior is $\Phi_{h+1}(u',w')$, where $u'$ is the updated suffix and $w'$ is obtained by normalising \eqref{eq: new weight construction}. Moreover, $w'$ is supported on $A_{h+1}(u')$.
\end{lemma}

The only remaining roadblock is discretisation: $h$ and $u$ are already discrete, but $w$ is not, since it is a probability vector of length (at most) $L$. The final step is a brute-force discretisation of the simplex. The proof of the following standard result is deferred to \Cref{sec: proof of discretize simplex}.

\begin{lemma} \label{lemma: discretize simplex}
    Let $d \geq 1$, $0 < \eta \leq 1$ and denote $\Delta_d$ the $(d-1)$-dimensional probability simplex in $\mathbb{R}^d$. There is a finite grid $G_{d,\eta} \subset \Delta_d$ such that for every $w \in \Delta_d$, there is a rounding point $g \in G_{d, \eta}$ with $\|w - g\|_1 \leq \eta$, and $|G_{d,\eta}| \leq (C/ \eta)^d$ for some universal constant $C$.
\end{lemma}

We now have all the elements to describe a finite planner. In the following, fix some $0<\eta \leq 1$, and let $G := G_{|\mathcal Q|, \eta}$. For each suffix, let $G_h(u):=\{g\in G:g_q=0\text{ for }q\notin A_h(u)\}$. A finite planner state is a triplet $m=(h,u,g)$ with $g\in G_h(u)$. Denote the collection of finite planner states at stage $h$ by $\mathcal M_h$. Note that $|\mathcal M_h|$ is at most $(AOS)^L\left(\frac{C}{\eta}\right)^{L}$, which will be quasipolynomial in the parameters if $(1/\eta)$ is $\mathrm{poly}$(parameters) and $L$ is $\mathrm{polylog}$(parameters).

For the planner, $\mathcal U_h$ contains all suffixes of the lengths in \cref{def: suffix} with $A_h(u)\neq\emptyset$, including those not reachable from $\mu_1$. Rounding is performed within $G_h(u)$, with the same accuracy and size bounds as in \cref{lemma: discretize simplex}; see \cref{sec: proof of discretize simplex}.

Suppose the planner is in state $m = (h,u, g)$. Given an action $a$ and an alive observation $z$, we describe how to arrive at the next state $m^+(m, a, z)$. The time and suffix indices are easily updated to $h^+$ and $u^+$. The planner computes the new weight vector $w^+:= w^+(m, a, z)$ via \cref{lemma: closure of actual belief}, taking the grid point $g$ as the current weight vector.\footnote{It can happen that  an observation $z$ that has 0 probability under $\beta_m := \Phi_h(u,g)$ is actually observed, since $\beta_m$ is not the ``true'' belief, but is an approximation maintained by the planner. If this happens, choose any admissible planner state at time $h+1$, not necessarily with the recorded suffix.} Then it rounds $w^+$ to $g^+\in G_{h+1}(u^+)$, preserving zeros outside $A_{h+1}(u^+)$ and incurring at most $\eta$ error in $\ell_1$. The new state is thus $m^+(m,a,z) := (h^+, u^+, g^+)$. 

\subsection{Finite-state planner}

With the finite planning model, we now describe how to compute a dynamic-programming based approximately optimal policy for the killed POMDP. Write $P(z\mid\beta,a)$ for the probability of the next observation $z$ under the killed dynamics, given current belief $\beta$ and action $a$, with the stage implicit. We first define $\widehat{V}_{H}(m) := 0$ for all $m \in \mathcal M_{H}$. For $h \in [1:H-1]$ and a finite planner state $m = (h,u,g)$, define
\begin{align}
    \widehat{V}_h(m)
    &:= \max_{a \in \mathcal A}
    \left\{
         \sum_{z = (y, \hat s) \in \mathcal Z}
        P(z \mid \beta_m,a) \left(r_{h+1}(y) + 
        \widehat{V}_{h+1}(m^+(m,a,z)) \right)
    \right\},
    \label{eq: Bellman value update}
    \\
    \widehat{a}_h(m)
    &\in \argmax_{a \in \mathcal A}
    \left\{
        \sum_{z = (y, \hat s) \in \mathcal Z}
        P(z \mid \beta_m,a) \left(r_{h+1}(y) +
        \widehat{V}_{h+1}(m^+(m,a,z))\right)
    \right\} .
    \label{eq: Bellman action selection}
\end{align}

The error analysis for this planner follows the two-step approach in \cite[Lemma 5.2--5.3]{gol22}. First, we bound the \emph{planning error}; i.e., we show that for each state $m$, the optimal remaining reward $V^*_h(\beta_m)$ from $\beta_m$ (the belief corresponding to $m$) is close to the anticipated reward $\hat V_h(m)$ recorded in the finite planner's table. The proof of the following result is in \Cref{app: proof of planning error}.

\begin{theorem} \label{thm: planning error bound}
Define $E_h^{\mathsf{plan}} := \max_{m \in \mathcal M_h} \left| V_h^*(\beta_m) - \widehat{V}_h(m)\right|$ to be the planning error. Then for each time $h$, $E_h^{\mathsf{plan}} \leq H^2\eta$.
\end{theorem}

Next, in \Cref{app: proof of execution error} we show \emph{execution error} is also small; i.e., the true reward is close to value recorded in the planner's table. 

\begin{theorem} \label{thm: execution error bound}
Let $E_h^{\mathsf{exec}} := \max_{m \in \mathcal M_h} \left|W_h(\beta_m, m) - \widehat{V}_h(m)\right|$ where $W_h(\beta, m)$ is the expected cumulative reward from time $h+1$ to $H$, when the hidden state distribution is $\beta$ and the finite planner has state $m$. Then for every time $h$, $E_h^{\mathsf{exec}} \leq H^2 \eta$.
\end{theorem}

Suppose we initialise the finite planner at time $1$ by $m_1 = (1, \emptyset, e_{\perp})$, indicating the history (suffix) at stage 1 is empty, and that no reveal has occurred yet, and let $\beta_1(s,q) = \mu_1(s) \mathbf{1}\{q = \perp\}$ be the initial belief. Then $V_L^* = V_1^*(\beta_1)$ and $V_L^{\widehat{\pi}} = W_1(\beta_1, m_1)$. Hence, the planning and execution error bounds from \cref{thm: planning error bound} and \cref{thm: execution error bound} show that 
\begin{align*}
    V_L^* - V_L^{\widehat{\pi}} \leq \left|V_1^*(\beta_{m_1}) - \widehat{V}_1(m_1)\right| + \left| \widehat{V}_{1}(m_1) - W_1(\beta_{m_1}, m_1)\right| \leq 2 H^2 \eta.
\end{align*}
In other words, the error of the planner is at most $2H^2 \eta$, and hence can be controlled by suitably choosing the parameter $\eta$.

\subsection{Planning for $\mathcal P$}

Observe that the finite state planner we have proposed for the killed-POMDP can be implemented as is for the original POMDP allowing for private randomness. Let's ignore the private randomness issue for now and analyse its runtime and error guarantees. Set
\begin{align}
    L := \min\left\{H, \left\lceil \frac{2}{\rho}\log\frac{4H^2}{\varepsilon}\right\rceil \right\} \qquad \text{and} \qquad \eta := \frac{\varepsilon}{8H^2}. \label{eq: choice of L}
\end{align}
If $L<H$, then the choice of $L$ gives
\begin{align*}
H^2(1-\rho/2)^L \leq H^2\exp(-\rho L/2) \leq \frac{\varepsilon}{4}.
\end{align*}
If $L=H$, there are only $H-1$ reveal flags, so a run of $L$ residual steps is impossible and the killing loss is zero. Thus, in either case, the killing loss is at most $\varepsilon/4$. Combining this with \Cref{thm: planning error bound} and \Cref{thm: execution error bound}, the proposed planner $\widehat\pi$ has error at most
\begin{align*}
V^* - V^{\widehat{\pi}} \leq \frac{\varepsilon}{4} + 2H^2\eta = \frac{\varepsilon}{2} < \varepsilon.
\end{align*}
This is the desired theoretical guarantee as in~\Cref{thm: main theorem}.

Substituting the choice of $\eta$, the total number of finite planner states in the killed POMDP $\widetilde{\mathcal P}$,  is at most 
\begin{align}
    \sum_{h=1}^H |\mathcal M_h| \leq H (AOS)^{L+1} \left(\frac{C_1H^2}{\varepsilon}\right)^{L+1} \leq \left(\frac{C AOSH}{\varepsilon}\right)^{3(L+1)}.
\end{align}
The last inequality holds for $0<\varepsilon\leq1$ and a suitable universal constant $C$.

To turn this state count into a runtime bound, observe that all operations required at each planner state have polynomial cost. For each pair $(h,u)$, the conditional distributions $(\phi^u_{h,q})_{q\in\mathcal Q}$ can be precomputed using at most $L$ updates per component, as in \cref{lemma: locality}. For each planner state $m$, action $a$, and observation $z$, the observation probability and updated weight vector are then obtained by finite sums over the physical states and last reveal times as in \Cref{lemma: closure of actual belief}. Rounding the weights and identifying the successor $m^+(m,a,z)$ also require only
polynomial time. Since $|\mathcal Z|=OS$, there are at most
\[
    AOS\sum_{h=1}^{H-1}|\mathcal M_h|
\]
state--action--observation combinations to process. Once these successors and probabilities have been computed, the number of operations needed by the Bellman  recursion can be bounded by
\[
    \operatorname{poly}\!\left(S,A,O,H,L,\log(1/\eta)\right)
    \sum_{h=1}^{H}|\mathcal M_h|.
\]
Thus, the bound on the number of states implies the desired quasipolynomial runtime bound. The derandomisation argument in \cref{app: derandomisation} can be done efficiently preserving the quasipolynomial bound. Lastly, the decoder channels can be found in polynomial time by solving linear programs (see \Cref{app: decoder is LP}). Hence, with $L$ as chosen in \eqref{eq: choice of L}, the overall runtime is the one stated in \Cref{thm: main theorem}. 

\begin{remark}
On the class of $\gamma$-$\ell_1$-observable POMDPs, the runtime of our algorithm differs from that of the algorithm from \citep{gol22}, although both are quasipolynomial. The runtime in that paper is $H(AO)^{\mathrm{O}(\gamma^{-4}\log(SH/\varepsilon))},$ which has better behaviour, for instance when $H$ is large.
\end{remark}

\section{Decodability from observability}
\label{sec: decodability from observability}

Our main result in this section is \Cref{thm: decodability from observability}, which shows that decodability is a weaker assumption than $\ell_1$ or $\ell_2$ observability. In particular, when $\gamma$ and $\alpha$ are constants, we immediately conclude quasipolynomial-time planning under $\ell_1$-observability (thus recovering the main result of \cite{gol22} in a different way) and under $\ell_2$-observability.

\begin{theorem}[Proved in \cref{app: decodability from observability}]\label{thm: decodability from observability}
Let $\alpha, \gamma \in (0,1)$. If $K \in \mathbb{R}^{O \times S}$ is $\gamma$-$\ell_1$-observable, then $K$ is $(\gamma^2/9)$-decodable. If $K$ is $\alpha$-$\ell_2$-observable, then $K$ is $(\alpha^2/6)$-decodable.
\end{theorem}

We only provide a brief proof sketch here.  A sufficient condition for constructing a good decoder is identifying $\mathbb{R}^S$-valued \textit{score-function} $g$ of the observation $Y$ that is unbiased and has small norm. Specifically, we want
\begin{align}
    \mathbb{E}[g(Y) | X = s] = e_s \quad \forall s \in \mathcal S, \quad \text{ and } \quad \| g(y) \|_2^2 \leq M \quad \forall y \in \mathcal O,
\end{align}
where $e_s$ is the standard basis vector. Writing $g_s(y)$ for the score of $s$ given $y$, when $g(y)\neq0$ we set the decoder $D$ to be $D(s|y) := \frac{g_s^2(y)}{\|g(y)\|_2^2}$. If $g(y)=0$, use any fixed probability distribution on $\mathcal S$. It is easy to show (see \Cref{lemma: score-function implies decoder}) that this decoder satisfies $\Pr(\hat X = s| X=s) \geq 1/M$ for all $s \in \mathcal S$. The problem thus reduces to identifying a good score function. 

Under $\alpha$-$\ell_2$-observability, $K$ is injective on $\mathbf{1}^\perp$, and its inverse $K^\dagger$ on $K(\mathbf{1}^\perp)$ has $2\to2$ operator norm at most $1/\alpha$. Composing this inverse with orthogonal projection onto $K(\mathbf{1}^\perp)$ gives an extension $K^*$ to $\mathbb R^O$ whose operator norm is also at most $1/\alpha$. For any fixed $p_0 \in \Delta(\mathcal S)$, it is easy to check that the score-function 
\begin{align}
    g(y) := p_0 + K^*(e_y - Kp_0). \label{eq: l2 score-function sketch}
\end{align}
satisfies unbiasedness and a norm-upper bound of $\lesssim 1/\alpha^2$, as desired. We defer the details to \Cref{app: decodability from observability}.

The same idea fails for $\gamma$-$\ell_1$-observability, because the projection step can blow up the operator norm $\|K^\dagger\|_{1 \to 1}$. The entire proof is a bit more technical, but has the same overall flavour of extending an inverse without blowing up its norm. We defer the details to \Cref{app: decodability from observability}.

\section{Conclusion} \label{sec: conclusion}

We proposed decodability as a natural and well-motivated assumption on observation channels under which quasipolynomial-time planning is possible in POMDPs. Decodability is especially appealing because it is easy to check and encompasses both $\ell_1$- and $\ell_2$-observability, the former of which can be hard to check. Unlike prior work, our proposed algorithm for decodable POMDPs does not rely on filter-stability arguments and thus presents a genuinely new way of representing the belief space. An interesting research question concerns memory requirements, as opposed to time complexity. Specifically, if the horizon and the sizes of the state, observation, and action spaces are at most $K$, how many bits of memory does a finite-state controller need to achieve a fixed additive approximation error? For fixed positive decodability and observability constants, respectively, our algorithm and that of \cite{gol22} give an $O((\log K)^2)$ upper bound on the memory retained during execution. Can we obtain matching lower bounds?

\section*{AI Disclosure}

OpenAI's ChatGPT suggested the main proof strategy for planning under $\ell_2$-observability. The authors recognised its applicability to the broader class of decodable POMDPs, developed this generalisation, and substantially simplified the proofs. The main text was written by the authors, which was then scanned using AI, and some wording was changed as per AI suggestions. The authors take responsibility for all mathematical claims and their presentation.

\section*{Acknowledgment}

Oliver Dobson thanks the Summer Research in Mathematics (SRIM) program for funding his internship in the summer of 2026. Varun Jog thanks OpenAI for providing access to ChatGPT through the ChatGPT for Academic Researchers programme. Varun Jog also acknowledges support from the Leverhulme Trust grant RPG-2025-226, ``Old Problems, New Perspectives: A Fresh Look at Classical Hypothesis Testing.'' Chun Hei Michael Shiu thanks the University of Cambridge for its hospitality during the visit in which this work was conducted, and acknowledges financial support from UBC’s Go Global International Learning Award.

\appendix
\crefalias{section}{appendix}

\section[Proofs for the Planning Algorithm]{Proofs for the planning algorithm} \label{app: proofs}

\subsection[Proof of Augmented Value Equivalence]{Proof of \cref{lemma: augmented value equivalence}} \label{sec: proof of augmented value eq}

Recall that allowing private randomisation does not change the optimal
value of a finite-horizon POMDP, so we may work with randomised policies in this proof. Any policy for the original POMDP $\mathcal P$ is automatically a valid policy for the augmented POMDP $\widehat{\mathcal P}$ that ignores $\widehat{X}_h$, hence $V^*_{\mathcal P} \leq V^*_{\widehat{\mathcal P}}$. Conversely, any policy $\pi_{\widehat{\mathcal P}}$ for the augmented POMDP $\widehat{\mathcal P}$ can be implemented as a policy $\pi_{\mathcal P}$ for the original POMDP $\mathcal P$ with additional private randomness. In particular, after seeing $Y_h = y$, the planner samples $\widehat{X}_h\sim D_h(\cdot |y)$ according to the optimal decoder, then follows the augmented policy $\pi_{\widehat{\mathcal P}}$. Therefore, the two classes have the same achievable values among randomised policies.

\subsection[Proof of Exact Reveal Decomposition]{Proof of \cref{lemma: exact-reveal decomposition}} \label{sec: proof of exact-reveal decomposition}

Recall that $d_{h,s} := \sum_y K_h(y|s) D_h(s|y)$ for each stage $h$ and hidden state $s$. Given the augmented channel $J_h$, define the reveal channel $R_h$ by 
\begin{align*}
    R_h((y, \widehat{s}) | s) = \begin{cases}
        \frac{J_h((y, s)| s)}{d_{h,s}}, & \text{if } \widehat{s} = s, \\
        0, & \text{if } \widehat{s} \neq s
    \end{cases}.
\end{align*}
For each fixed hidden state $s$, 
\begin{align*}
    \sum_{y,\widehat{s}} R_h((y,\widehat{s}) | s) = \sum_{y} R_h((y,s)| s) = \frac{1}{d_{h,s}} \sum_y K_h(y|s) D_h(s|y) = 1.
\end{align*}
Hence $R_h$ is stochastic, and it is clear from the construction that $R_h((y, \widehat{s}) | s) > 0 \implies \widehat{s} = s$.

Next, we define the residual channel by $N_h := \frac{J_h - cR_h}{1-c}$. If $\widehat{s} \neq s$, 
\begin{align*}
    N_h((y,\widehat{s})| s) = \frac{J_h((y,\widehat{s})| s)}{1-c} \geq 0.
\end{align*}
If $\widehat{s} = s$, then 
\begin{align*}
   N_h((y,s)|s) & = \frac{J_h((y,s)| s) - cR_h((y,s)|s)}{1-c} \\
   & = \frac{1}{1-c} \left( J_h((y,s)|s) - \frac{c}{d_{h,s}} J_h((y,s)|s)\right) \\
   & = \frac{J_h((y,s)|s)}{1-c}\left(1 - \frac{c}{d_{h,s}}\right) \geq 0
\end{align*}
since  $c = \rho/2 \leq \rho \leq d_{h,s}$ for all $h,s$. Hence, $N_h$ is also entry-wise non-negative. On the other hand,
\begin{align*}
    \sum_{y,\widehat{s}} N_h((y,\widehat{s}) |s) & = \frac{1}{1-c}\left(\sum_{y, \widehat{s} \neq s} J_h((y,\widehat{s})| s) + \sum_y J_h((y,s)| s) \left(1 - \frac{c}{d_{h,s}}\right) \right) \\
    & = \frac{1}{1-c}\left(\sum_{y, \widehat{s}} J_h((y, \widehat{s})|s) - \frac{c}{d_{h,s}} \sum_y J_h((y,s)|s)\right) \\
    & = \frac{1}{1-c}\left(1 - \frac{c}{d_{h,s}} \sum_y K_h(y|s) D_h(s|y)\right) \\
    & = \frac{1}{1-c} \left(1-\frac{c}{d_{h,s}} \cdot d_{h,s}\right) = 1.
\end{align*}
Hence, $N_h$ is stochastic, and it follows from the construction of $R_h$ and $N_h$ that $J_h = cR_h + (1-c) N_h$.

\subsection[Proof of \Cref{lemma: killing loss}]{Proof of \Cref{lemma: killing loss}}\label{appendix: killing loss}

We first apply the coupling argument. Couple the augmented POMDP with the killed POMDP using the same hidden states, actions, transitions, latent variables $F_h$, and alive augmented observations until the killing event. By the coupling, two trajectories are identical on $E_L^c$. On $E_L$, their total rewards can differ by at most $H$ because all rewards are upper bounded by 1. Hence, for any policy $\pi$ for the augmented POMDP, if we use the same action rule for the killed POMDP on alive histories, we have
    \begin{align*}
        V_L^\pi \geq V^\pi - H \Pr(E_L) \geq V^\pi - H^2(1-c)^L.
    \end{align*}
    Applying this inequality to an optimal policy $\pi^*$ for the augmented POMDP, then maximizing over the killed POMDP side yields 
    \begin{align*}
        V_L^* \geq V_L^{\pi^*} \geq V^{\pi^*} - H^2(1-c)^L = V^* - H^2(1-c)^{L}.
    \end{align*}

    On the other hand, given a policy $\pi_L$ for the killed POMDP, define a policy $\pi$ for the augmented POMDP that agrees with $\pi_L$ on every alive history $\mathsf{h}_h$ before the process is killed; and behaves arbitrarily on histories that have already produced a cemetery observation $\dagger$. Under the coupling, the two POMDPs receive identical rewards up to killing. After killing, the killed POMDP obtains zero reward, whereas the augmented/original POMDP continues to receive non-negative rewards. Thus the realised total reward in the augmented POMDP is at least that in the killed POMDP on every coupled trajectory. Taking expectations gives $V^\pi \geq V_L^{\pi_L}$. Combining with the assumption $V_L^{\pi_L} \geq V_L^* - \eta_0$ yields 
    \begin{align*}
        V^\pi \geq V_L^{\pi_L} \geq V_L^* - \eta_0 \geq V^* - H^2(1-c)^L - \eta_0.
    \end{align*}

\subsection[Proof of \Cref{lemma: locality}]{Proof of \Cref{lemma: locality}}\label{appendix: locality}

    We first discuss the trivial case when $q = \perp$, in which no reveal has occurred yet. An alive no-reveal path contains fewer than $L$ residual observations, so the entire observation and action history from the initial belief $\widetilde{\mu}_1$ is contained in $u_h$. The conditional state belief given $Q_h=\perp$ is obtained from $\mu_1$ using transitions and residual observation updates, so it is determined by $(h,u_h,q)$. Its posterior weight additionally requires normalisation by the likelihood of the full observed history under the killed model.

    Next, suppose $q \in \{0,1,\ldots, L-1\}$. Then the most recent hidden state exact-reveal occurred at time $\tau := h-q$, and the observation at that time is $\widetilde{Z}_\tau = (Y_\tau, \widehat{X}_\tau)$. Since it is an exact-reveal, $F_\tau = 1$ and hence $\widehat{X}_\tau = X_\tau$ with probability one. Therefore, the conditional distribution at time $\tau$ is the point mass $\delta_{\widehat{X}_\tau}$. 

    Since $\tau=h-q$ is the most recent reveal, $F_{\tau+1}=\cdots=F_h=0$. Starting from $\delta_{\widehat{X}_\tau}$, apply transitions $T_\tau,\ldots,T_{h-1}$ with the recorded actions and residual observation updates $N_{\tau+1},\ldots,N_h$. By the Markov property, no information before $\tau$ is needed once $X_\tau$ is fixed. Since $h-\tau=q<L$, we have $\tau\geq h-L+1$, so the reveal observation and all subsequent actions and observations lie in $u_h$. Thus the conditional state belief is determined by $(h,u_h,q)$.

\subsection[Proof of \Cref{lemma: closure of actual belief}]{Proof of \Cref{lemma: closure of actual belief}}\label{appendix: closure of actual belief}
Fix $\beta=\Phi_h(u,w)$ with $w$ supported on $A_h(u)$, and write $w_{h,q}=w_q$ and $\phi_{h,q}=\phi_{h,q}^u$. All probabilities below are computed from this representative belief.

On a reveal branch, $R_{h+1}(\widetilde{z} | s') > 0$ implies $s' = \widehat{s}(\widetilde{z})$, where $\widehat{s}(\widetilde{z})$ denotes the second coordinate for an alive observation $\widetilde{z} = (y, \widehat{s})$. So the conditional distribution is a point mass on $\widehat{s}(z)$. Define
    \begin{align*}
        \psi_q(s') = \sum_s T_h(s'|s,a) \cdot \phi_{h,q}(s) 
    \end{align*}
    for $q \in \mathcal Q$. Applying Bayes' rule to the reveal and surviving residual branches gives the unnormalised weights for stage $h+1$ and alive observation $\widetilde{z} \in \mathcal Z$:
    \begin{align} \label{eq: new weight construction}
        \overline{w}_{h+1,q'}(\widetilde{z}) = \begin{cases}
            c \sum_{q \in \mathcal Q} w_{h,q} \sum_{s'} R_{h+1}(\widetilde{z}|s') \cdot \psi_q(s'), & \text{if } q'= 0, \\
            (1-c) w_{h,q} \sum_{s'} N_{h+1} (\widetilde{z} | s') \cdot \psi_q(s'), & \text{if } q' = q+1 \in \{1,2, \ldots, L-1\}, \\
            (1-c) w_{h,\perp}\mathbf 1\{h<L\} \sum_{s'} N_{h+1}(\widetilde z|s') \cdot \psi_{\perp}(s'), & \text{if } q'=\perp.
        \end{cases} 
    \end{align}
    The vector $\overline{w}_{h+1,q'}$ is not a probability vector since we have only defined it for alive observations $\widetilde{z}$, so
    \begin{align*}
        \sum_{\widetilde z\text{ alive}}\sum_{q'\in\mathcal Q}\overline w_{h+1,q'}(\widetilde z)
        =\sum_{\widetilde z\text{ alive}}P(\widetilde z\mid\beta,a)
        =1-P(\dagger\mid\beta,a).
    \end{align*}
    Define the next-stage killing probability
    \begin{align*}
        p_{h+1,\dagger}:=P(\dagger\mid\beta,a)=(1-c)\bigl(w_{h,L-1}+\mathbf 1\{h\geq L\}w_{h,\perp}\bigr).
    \end{align*}
For an alive observation $\widetilde{z}$, define $\Lambda_h(\widetilde{z}) := \sum_{q' \in \mathcal Q} \overline{w}_{h+1,q'}(\widetilde{z})$. If an alive observation $\widetilde{z}$ is observed, then the survival of the process at stage $h+1$ is known. For $\widetilde{z}$ such that $\Lambda_h(\widetilde{z}) > 0$, we normalize
    \begin{align}
        w'_{q'}(\widetilde z)=\frac{\overline w_{h+1,q'}(\widetilde z)}{\Lambda_h(\widetilde z)}=\Pr_\beta(Q_{h+1}=q'\mid A_h=a,\widetilde Z_{h+1}=\widetilde z).\label{eq: posterior update rule}
    \end{align}

It remains to show that the conditional state beliefs agree with those defined by $u'$. Consider an age with positive posterior weight. If a reveal occurs at time $h+1$, the state is $\widehat s(\widetilde z)$, so age $0$ is admissible and its conditional belief is $\phi_{h+1,0}^{u'}$. Otherwise, a numerical age $q$ increases to $q+1$. The conditional belief is obtained by applying one more transition and residual-observation update to the calculation starting at time $h-q$. Since $q+1\leq L-1$, $u'$ contains all the observations and actions needed for this calculation, which defines $\phi_{h+1,q+1}^{u'}$. For $q=\perp$, survival requires $h<L$, so $u'$ contains the full history and the calculation starts from $\mu_1$. In each case the calculation has positive likelihood. Thus $w'$ is supported on $A_{h+1}(u')$, and the posterior is $\Phi_{h+1}(u',w')$.

\subsection[Proof of Discretize Simplex]{Proof of \cref{lemma: discretize simplex}} \label{sec: proof of discretize simplex}

Let $n = \left\lceil \frac{2d}{\eta}\right\rceil$ and define
\begin{align*}
    G_{d,\eta} : =\left\{ \left(\frac{k_1}{n}, \frac{k_2}{n},\ldots, \frac{k_d}{n}\right): 0 \leq k_i \in \mathbb{Z}, \sum_{i=1}^d k_i = n\right\}.
\end{align*}
Given a probability vector $w \in \Delta_d$, round down the first $d-1$ coordinates to the nearest multiples of $\frac{1}{n}$ and put the remaining mass in the last coordinate.
Let $g$ be the resulting grid point. Each of the first $d-1$ coordinates loses at most $1/n$, and the last coordinate gains their total loss. Hence
\begin{align*}
    \|w-g\|_1 \leq \frac{2(d-1)}{n} \leq \eta.
\end{align*}
The number of grid points in $G_{d, \eta}$ is 
\begin{align*}
    |G_{d,\eta}| = \binom{n+d-1}{d-1} \leq \binom{n+d}{d} \leq \left(\frac{e(n+d)}{d}\right)^d \leq \left(\frac{4e}{\eta}\right)^d,
\end{align*}
where the last inequality uses $n\leq 2d/\eta+1$, $d\geq1$, and $0<\eta\leq1$. Thus we may take $C=4e$.

For a nonempty admissible set $A_h(u)$ of size $k\leq d$, use the same lattice grid restricted to that simplex face. Round down all but one admissible coordinate to multiples of $1/n$, put the remaining mass in the last admissible coordinate, and keep all other coordinates zero. The result belongs to $G_h(u)$ and has $\ell_1$ error at most $2(k-1)/n\leq\eta$, since $k\leq d$ and $n\geq2d/\eta$. For $d=1$, the error is zero. Restricting the grid does not increase its size.

\subsection[Proof of Planning Error]{Proof of \cref{thm: planning error bound}} \label{app: proof of planning error}

Before proving this result, we state and prove two simple lemmas.

\begin{lemma}\label{lemma: belief isometry}
    For a fixed stage-suffix pair $(h,u)$, for any $w, w' \in \Delta(\mathcal Q)$, the belief $\Phi_h(u, \cdot)$ is an isometry in the 1-norm, i.e.,
    \begin{align}
        \|\Phi_h(u,w) - \Phi_h(u,w')\|_1 = \|w - w'\|_1. \label{eq: belief isometry}
    \end{align}
\end{lemma}
\begin{proof}
Note that most recent reveal values are on disjoint coordinates of the augmented hidden state. Therefore,
    \begin{align*}
        \|\Phi_h(u,w) - \Phi_h(u,w')\|_1 & = \sum_{q \in \mathcal Q} \sum_s |w_q \cdot \phi_{h,q}^u(s) - w_q' \cdot \phi_{h,q}^u(s)| \\
        & = \sum_{q \in \mathcal Q} |w_q - w_q'| \sum_{s} \phi_{h,q}^u(s) \\
        & = \|w - w'\|_1
    \end{align*}
    since $\phi_{h,q}^u$ is a conditional distribution.
\end{proof}

\begin{lemma}\label{lemma: value are Lipschitz}
    Let $h$ be a stage and $\pi$ be a policy, and let $\beta, \beta'$ be two beliefs over the augmented hidden states. Then
    \begin{align}
        |V_h^\pi(\beta) - V_h^\pi(\beta')| \leq H\|\beta - \beta'\|_1. \label{eq: V is H-Lipschitz}
    \end{align}
    The same Lipschitz condition holds for the optimal value $V_h^*$.
\end{lemma}

\begin{proof}
  Let $v_h^\pi(\xi)$ be the expected cumulative reward from time $h+1$ to $H$ given that the augmented hidden state is $\xi$. It is clear that $0 \leq v_h^\pi(\xi) \leq H$ for $h$ and policy $\pi$ since $r_h \in [0,1]$. It follows by triangle inequality that 
\begin{align*}
    |V_h^\pi(\beta) - V_h^\pi(\beta')| & = \left|\sum_{\xi} (\beta(\xi) - \beta'(\xi)) v_h^\pi(\xi)\right| \leq H \|\beta - \beta'\|_1.
\end{align*}
Taking supremum over all policies on both sides gives the same Lipschitz bound for $V_h^*$.  
\end{proof}

We now proceed by backward induction. At time $H$, both $V_{H}^*(\beta_m)$ and $\widehat{V}_{H}(m)$ are zero, so $E_{H}^\mathrm{plan} = 0$.

Fix $h<H$ and a finite planner state $m \in \mathcal M_h$. For each action $a$, define
\begin{align*}
    Q_h(\beta_m, a)
    &:= \sum_{z=(y,\widehat s)\in\mathcal Z}
    P(z|\beta_m,a)
    \left[r_{h+1}(y) + V_{h+1}^*(\beta_{m,a}^+(z))\right], \\
    \widehat{Q}_h(m,a)
    &:= \sum_{z=(y,\widehat s)\in\mathcal Z}
    P(z|\beta_m,a)
    \left[r_{h+1}(y) + \widehat{V}_{h+1}(m^+(m,a,z))\right],
\end{align*}
where $\beta_{m,a}^+(z)$ is the exact posterior after taking action $a$ and observing $z$. Zero-probability terms are omitted from the sums. In the following, we write $m^+ = m^+(m,a,z)$. For each action $a \in \mathcal A$ and observation $z$ with $P(z|\beta_m,a)>0$, admissible closure (\cref{lemma: closure of actual belief}), rounding within $G_{h+1}(u^+)$, and the isometry in \cref{eq: belief isometry} give
\begin{align}
    \|\beta_{m,a}^+(z) - \beta_{m^+}\|_1 \leq \eta. \label{eq: isometry and grid}
\end{align}
Hence, by the Lipschitz condition in \cref{lemma: value are Lipschitz},
\begin{align*}
    \left|V_{h+1}^*(\beta_{m,a}^+(z)) - \widehat{V}_{h+1}(m^+)\right|
    &\leq \left|V_{h+1}^*(\beta_{m,a}^+(z)) - V_{h+1}^*(\beta_{m^+})\right|
    + \left|V_{h+1}^*(\beta_{m^+}) - \widehat{V}_{h+1}(m^+)\right| \\
    &\leq H \eta + E_{h+1}^\mathsf{plan}.
\end{align*}

By the definition of exact and rounded Bellman recursions,
\begin{align*}
    V_h^*(\beta_m) = \max_{a \in \mathcal A} Q_h(\beta_m, a)
    \qquad \text{and} \qquad
    \widehat{V}_h(m) = \max_{a\in\mathcal A} \widehat{Q}_h(m,a).
\end{align*}
Therefore,
\begin{align*}
    \left| V_h^*(\beta_m) - \widehat{V}_h(m)\right|
    &= \left| \max_{a \in\mathcal A} Q_h(\beta_m, a)
    - \max_{a'\in\mathcal A} \widehat{Q}_h(m,a')\right| \\
    &\leq \max_{a \in \mathcal A}
    \left| Q_h(\beta_m, a) - \widehat{Q}_h(m,a)\right| \\
    &= \max_{a \in \mathcal A}
    \left|\sum_z P(z|\beta_m, a)
    \left(V_{h+1}^*(\beta_{m,a}^+(z))
    - \widehat{V}_{h+1}(m^+(m,a,z))\right)\right| \\
    &\leq \max_{a \in \mathcal A}
    \sum_z P(z|\beta_m, a)
    \left|V_{h+1}^*(\beta_{m,a}^+(z))
    - \widehat{V}_{h+1}(m^+(m,a,z))\right| \\
    &\leq H \eta + E_{h+1}^\mathsf{plan}.
\end{align*}
Since this is true for all $m \in \mathcal M_h$, we obtain
\begin{align*}
    E_h^{\mathsf{plan}}
    = \max_{m \in \mathcal M_h}
    \left| V_h^*(\beta_m) - \widehat{V}_h(m)\right|
    \leq H \eta + E_{h+1}^\mathsf{plan}.
\end{align*}
Iterating from $E_{H}^\mathsf{plan} = 0$ yields
$E_h^\mathsf{plan} \leq H(H-h)\eta \leq H^2\eta$.

\subsection[Proof of Execution error bound]{Proof of \cref{thm: execution error bound}} \label{app: proof of execution error}

We prove by backward induction. At stage $H$, both the finite planner's continuation value and the computed value from the Bellman table are zero, so $E_{H}^\mathsf{exec} = 0$.

Fix $h<H$ and a finite planner state $m \in \mathcal M_h$, and write $a = \widehat{a}_h(m)$ to be the action selected by the finite planner. As in \cref{app: proof of planning error}, let $\beta_{m,a}^+(z)$ be the exact posterior obtained from $\beta_m$ after taking action $a$ and observing a regular observation $z$ with $P(z|\beta_m,a)>0$. Also, we write $m^+ = m^+(m,a,z)$. Zero-probability terms are omitted from the sums below.

For a fixed memory state $m^+$, the finite planner follows a fixed policy, and thus $W_{h+1}(\cdot, m^+)$ also satisfies the Lipschitz bound. Hence, using \cref{eq: isometry and grid}, we have
\begin{align*}
    \left|W_{h+1}(\beta_{m,a}^+(z), m^+) - \widehat{V}_{h+1}(m^+)\right|
    &\leq \left|W_{h+1}(\beta_{m,a}^+(z), m^+) - W_{h+1}(\beta_{m^+}, m^+)\right| \\
    &\quad + \left|W_{h+1}(\beta_{m^+}, m^+) - \widehat{V}_{h+1}(m^+)\right| \\
    &\leq H \eta + E_{h+1}^\mathsf{exec}.
\end{align*}
Conditioning on the next-stage observation, the finite planner's value can be expressed as
\begin{align*}
    W_h(\beta_m,m)
    = \sum_{z=(y,\widehat s)\in\mathcal Z}
    P(z|\beta_m,a)
    \left[r_{h+1}(y) + W_{h+1}(\beta_{m,a}^+(z),m^+)\right].
\end{align*}
On the other hand, $a = \widehat{a}_h(m)$ is a maximiser in the rounded Bellman recursion, so
\begin{align*}
    \widehat{V}_h(m)
    = \sum_{z=(y,\widehat s)\in\mathcal Z}
    P(z|\beta_m,a)
    \left[r_{h+1}(y) + \widehat{V}_{h+1}(m^+)\right].
\end{align*}
Therefore,
\begin{align*}
    \left|W_h(\beta_m,m) - \widehat{V}_h(m)\right|
    &= \left|\sum_z P(z|\beta_m,a)
    \left(W_{h+1}(\beta_{m,a}^+(z),m^+)
    - \widehat{V}_{h+1}(m^+)\right)\right| \\
    &\leq \sum_z P(z|\beta_m,a)
    \left|W_{h+1}(\beta_{m,a}^+(z),m^+)
    - \widehat{V}_{h+1}(m^+)\right| \\
    &\leq H \eta + E_{h+1}^\mathsf{exec}.
\end{align*}
Since this is true for all $m \in \mathcal M_h$, we obtain
\begin{align*}
    E_h^\mathsf{exec}
    = \max_{m \in \mathcal M_h}
    \left|W_h(\beta_m,m) - \widehat{V}_h(m)\right|
    \leq H\eta + E_{h+1}^\mathsf{exec}.
\end{align*}
Iterating from $E_{H}^\mathsf{exec} = 0$ yields
$E_h^\mathsf{exec} \leq H(H-h)\eta \leq H^2\eta$.

\subsection{Finding decoders in polynomial-time} \label{app: decoder is LP}

Let $K \in \mathbb{R}^{O \times S}$ be an observation kernel. Let $\mathcal D$ be the set of all column-stochastic matrices in $\mathbb R^{S\times O}$, representing decoders from $\mathcal O$ to $\mathcal S$. Recall that 
\begin{align*}
    \rho(K) := \max_{D \in \mathcal D} \min_{s \in \mathcal S} d_s(D) = \max_{D \in \mathcal D} \min_{s \in \mathcal S} \sum_y K(y|s) D(s|y).
\end{align*}
Then $\rho(K)$ is the optimal value of the following linear program:
\begin{equation}
\label{eq: LP of optimal decoder}
\begin{aligned}
    \max \quad & \rho \\
    \text{subject to} \quad
    & \sum_y K(y \mid s) D(s \mid y) \geq \rho
    && \text{for all } s, \\
    & \sum_s D(s \mid y) = 1
    && \text{for all } y, \\
    & D(s \mid y) \geq 0
    && \text{for all } s,y.
\end{aligned}
\end{equation}
Since the above optimization problem has a polynomial number of variables and constraints, it is computable in polynomial time  in $S, O$, and the input bit complexity of $K$.

\subsection{Efficient derandomisation} \label{app: derandomisation}
In the following, we show that the derandomisation can be done in quasipolynomial time. More specifically, it can be done in time polynomial in $S, O, H$, and $|\Pi|$, where $\Pi$ is the randomised finite-state controller and $|\Pi|$ is its description size. The decoding table contains one entry for each stage $h \in \{2,\ldots,H\}$ and observation $y\in\mathcal O$, giving $(H-1)O$ entries in total. We fill these entries sequentially. For each entry, we try every decoded state in the support of $D_h(\cdot\mid y)$, keeping previously selected entries fixed and leaving all unfilled entries randomized according to their original decoder distributions. We compute the expected total reward of each candidate controller and select a candidate with the highest value.

Each candidate is evaluated by forward dynamic programming on
hidden-state memory pairs. At stage $h$, we maintain the probability table $\Pr(X_h=s,m_h=m)$ and propagate it using the transition and observation kernels of $\mathcal P$, together with the existing action and memory-update tables. Since rewards depend only on observations, we compute the expected reward at each stage by weighting each observation's reward by its probability, and sum these contributions over the horizon. Thus, each evaluation uses the existing controller tables without repeating the Bellman optimization.

Let $M$ be the maximum number of controller memory states at any stage, clearly $M\leq |\Pi|$. The probability table at each stage has at most $SM$ entries. A straightforward implementation loops over the current hidden state, current memory, next hidden state, next observation, and decoded state, while obtaining the action and next memory by table lookup. Consequently, each forward DP evaluation takes polynomial number of operations in $S, O, H$, and $M$. Since there are at most $HO$ decoding-table entries and at most
$S$ candidates per entry, the extraction can be done in time polynomial in $S$, $O$, $H$, and $|\Pi|$.

Finally, independently sampling all decoding-table entries in advance induces the same trajectory distribution as online decoder sampling, because each time-indexed entry is used at most once along a trajectory. At each derandomisation step, the current expected value is a weighted average of the candidate values for the next entry. Selecting a maximizing candidate therefore cannot decrease the expected value. After all entries have been fixed, the resulting controller $\widehat\pi$ is deterministic and satisfies $V_{\mathcal P}^{\widehat\pi}\geq V_{\mathcal P}^{\Pi}$.

\section{Proofs for \Cref{sec: decodability from observability}}
\label{app: decodability from observability}

Our framework seeks an $\mathbb{R}^S$-valued \textit{score-function} $g$ of the observation $Y$ such that 
\begin{align}
    \mathbb{E}[g(Y) | X = s] = e_s \qquad \text{for every } s \in \mathcal S, \label{eq: unbias score-function}
\end{align}
where $e_s$ denotes the standard basis vector corresponding to state $s$. Equivalently, the $t$-th coordinate of $g$ satisfies
\begin{align*}
    \mathbb{E}[g_t(Y)| X=s] = \mathbf{1}\{s=t\}.
\end{align*}
Thus, the expected score of the true state is 1, while the expected score of every other state is 0. The following lemma states that we can convert a bounded unbiased score function into a decoder with guaranteed performance.

\begin{lemma} \label{lemma: score-function implies decoder}
    Let $M \geq 1$, and suppose $g : \mathcal O \to \mathbb{R}^S$ is a score-function that satisfies \eqref{eq: unbias score-function} and
    \begin{align}
        \|g(y)\|_2^2 = \sum_{s\in\mathcal S} g_s^2(y) \leq M \qquad \text{for every } y \in \mathcal O.  \label{eq: score-function bounded norm}
     \end{align}
     Then there exists a decoder with worst-state success-decoding probability at least $1/M$ for every hidden state.
\end{lemma}

\begin{proof}
    For $g(y) \neq 0$, define the decoder as 
    \begin{align*}
        D(s|y) := \frac{g_s^2(y)}{\|g(y)\|_2^2}.
    \end{align*}
    If $g(y) = 0$, use any fixed auxiliary probability distribution on $\mathcal S$. This is a decoder that depends only on the observation $y$, not on the hidden state $s$. \eqref{eq: score-function bounded norm} implies that 
    \begin{align*}
        D(s|y) \geq \frac{g_s^2(y)}{M}
    \end{align*}
    for every $s \in \mathcal S$ and $y \in \mathcal O$, and it holds immediately when $g(y) = 0$. Consequently, 
    \begin{align*}
        \Pr(\widehat{X} = s | X = s) & = \mathbb{E}[ D(s|Y) | X =s ] \\
        & \geq \mathbb{E}\left[ \frac{g_s^2(Y)}{M} | X =s\right]\\
        & \geq \frac{1}{M} \left( \mathbb{E}[g_s(Y) | X=s]\right)^2 \\
        & = \frac{1}{M}
    \end{align*}
    for every hidden state $s\in \mathcal S$, where we used Jensen's inequality in the last inequality.
\end{proof}

Therefore, it remains to construct unbiased score-functions with a controlled norm. Observe that 
\begin{align*}
    \mathbb{E}[e_Y | X=s] = Ke_s,
\end{align*}
which suggests to interpret the score-function as recovering an unbiased estimate of $e_s$ by inverting $K$. Let $\mathbf{1}_S^\perp$ and $\mathbf{1}_O^\perp$ denote the zero-sum subspaces of $\mathbb{R}^S$ and $\mathbb{R}^O$, respectively. Since $K$ is an observation kernel, $K(\mathbf{1}_S^\perp) \subseteq \mathbf{1}_O^\perp$. On the other hand, observability implies $K$ is injective on $\mathbf{1}_S^\perp$, and hence invertible on $K(\mathbf{1}_S^\perp)$.

The idea is to extend this inverse, or suitable scalar functionals of it while controlling the resulting scores. Under $\ell_2$-observability, a natural orthogonal projection gives a bounded linear extension. However, the same approach fails in the $\ell_1$-case. Instead, we use a characterization with random sign vectors.

\subsection{\texorpdfstring{$\ell_2$}{l2}-Observability}\label{sec: l2-implies-decodability}

Let $K$ be an observation kernel that satisfies $\alpha$-$\ell_2$-observability. Then the pseudo-inverse $K^\dagger$ of $K$ on $K(\mathbf{1}_S^\perp)$ has operator norm at most $1/\alpha$. Composing this inverse with the orthogonal projection onto $K(\mathbf{1}_S^\perp)$ gives a linear map 
\begin{align*}
    K^* := K^\dagger \circ  \Pi_{K(\mathbf{1}_S^\perp)} : \mathbf{1}_O^\perp \to \mathbf{1}_S^\perp 
\end{align*}
that satisfies $K^*K v = v$ for every $v \in \mathbf{1}_S^\perp$ and $\|K^*\|_{2\to2} \leq \|K^\dagger\|_{2 \to 2} \|\Pi_{K(\mathbf{1}_S^\perp)}\|_{2 \to 2} \leq 1/\alpha$.

Let $p_0 \in \Delta(\mathcal S)$ be a reference distribution, and define
\begin{align}
    g(y) := p_0 + K^*(e_y - Kp_0). \label{eq: l2 score-function}
\end{align}
Then for every hidden state $s\in\mathcal S$,
\begin{align*}
    \mathbb{E}[g(Y) | X=s]& = \mathbb{E}[p_0 + K^*(e_Y - Kp_0)|X=s]\\
    & = p_0 + K^*(Ke_s - Kp_0) \\
    & = p_0 + K^*K(e_s - p_0) \\
    & = e_s,
\end{align*}
since $e_s - p_0 \in \mathbf{1}_S^\perp$. Thus $g$ is indeed unbiased.

The score function is also independent of the reference distribution, as
\begin{align*}
    [p_0 + K^*(e_y - Kp_0)] - [p_1 + K^*(e_y - Kp_1)] = (p_0 - p_1) - K^*K(p_0 -p_1) = 0
\end{align*}
for any $p_0, p_1 \in \Delta(\mathcal S)$. Therefore, we may use $p_0 = e_s$ and for every observation $y \in \mathcal O$,
\begin{align*}
    \|g(y)\|_2 & = \left\|e_s + K^*(e_y - Ke_s) \right\|_2 \\
    & \leq \|e_s\|_2 + \|K^*\|_{2 \to 2} \|e_y - Ke_s\|_2 \\
    & \leq 1 +\frac{\sqrt{2}}{\alpha},
\end{align*}
where we use the fact that the $\ell_2$-distance between two probability vectors is at most $\sqrt{2}$. Apply \cref{lemma: score-function implies decoder} with $M = \left(1 + \frac{\sqrt{2}}{\alpha} \right)^2$ yields a decoder with worst-case success-decoding probability
\begin{align*}
    \Pr(\widehat{X} = s | X =s) \geq \frac{\alpha^2}{(\alpha+\sqrt{2})^2} \geq \frac{\alpha^2}{6} \qquad \text{for all } s \in\mathcal S.
\end{align*}

\subsection{\texorpdfstring{$\ell_1$}{l1}-Observability}\label{sec: l1-implies-decodability}

We now turn to constructing a score function under $\ell_1$-observability. Note that the extension of a pseudo-inverse with an orthogonal projection need not be contractive in the $\ell_1$ case, so the construction used in \cref{sec: l2-implies-decodability} does not directly preserve the desired norm bound. Instead of extending the vector-valued inverse, we extend scalar functionals of the inverse, then combine them into an unbiased score-function.

The choice of these functionals is motivated by the characterization
\begin{align}
    \|v\|_1 = \max_{\sigma \in \{-1,+1\}^S} \langle \sigma, v\rangle,
\end{align}
where $\langle\cdot, \cdot\rangle$ denotes the standard inner product in $\mathbb{R}^S$. Thus, sign vectors provide a natural family of scalar measurements controlled by the $\ell_1$-norm. Also, they allow us to reconstruct the vector $v$ by averaging.

Let $K$ be an observation kernel that satisfies $\gamma$-$\ell_1$-observability. For each random sign vector $\sigma \in \{-1, +1\}^S$, define a linear functional $l_\sigma$ on $K(\mathbf{1}_S^\perp)$ by
\begin{align*}
    l_\sigma(Kv) := \langle \sigma, v\rangle 
\end{align*}
for $v \in \mathbf{1}_S^\perp$. Again, this is well-defined since observability implies that $K$ is injective on $\mathbf{1}_S^\perp$. Moreover, note that
\begin{align*}
    |l_\sigma(Kv) | = |\langle \sigma, v \rangle| \leq \|v\|_1 \leq \frac{1}{\gamma}\|Kv\|_1.
\end{align*}
So by Hahn-Banach theorem, there exists some $a_\sigma \in \mathbb{R}^O$ such that
\begin{align*}
    \|a_\sigma\|_\infty \leq \frac{1}{\gamma} \qquad \text{and} \qquad \langle a_\sigma, Kv\rangle = \langle \sigma, v\rangle \qquad \text{for all } v\in \mathbf{1}_S^\perp.
\end{align*}
In other words, $a_\sigma$ is an extension of $l_\sigma$ while the dual-norm is controlled. Let $p_0 \in \Delta(\mathcal S)$ be a reference distribution, and define the scalar signed score as 
\begin{align*}
    c_\sigma(y) := \langle \sigma, p_0\rangle + \langle a_\sigma, e_y - Kp_0\rangle.
\end{align*}
Then for every hidden state $s$,
\begin{align*}
    \mathbb{E}[c_\sigma(Y)|X=s]& = \mathbb{E}[\langle \sigma, p_0\rangle + \langle a_\sigma, Ke_s - Kp_0\rangle | X=s]\\
    & = \langle \sigma,p_0\rangle + \langle \sigma, e_s - p_0\rangle \\
    & = \langle \sigma, e_s\rangle \\
    & = \sigma_s.
\end{align*}
Thus, one can view $c_\sigma(Y)$ as an unbiased estimate of the $s$-th coordinate in the sign vector $\sigma$. Furthermore,
\begin{align*}
    |c_\sigma(y)| & \leq |\langle \sigma, p_0\rangle| + \|a_\sigma\|_\infty \|e_y - Kp_0\|_1 \leq 1 + \frac{2}{\gamma},
\end{align*}
where we used the fact that the $\ell_1$-distance between two probability vectors is at most $2$.

Next, we combine these scalar scores $c_\sigma(Y)$ to reconstruct the vector-valued score-function. Observe that if $\sigma \sim \mathrm{Unif}\left(\{-1,+1\}^S\right)$ and its coordinates are independent uniform signs, then $\mathbb{E}_\sigma[\sigma_s \sigma_t] = \mathbf{1}\{s=t\}$ for all $s,t \in\mathcal S$. Then every $v \in \mathbb{R}^S$ satisfies the reconstruction formula $v = \mathbb{E}_\sigma[\sigma \langle \sigma, v\rangle]$. Therefore, we define the score-function as 
\begin{align*}
    g(y) := \mathbb{E}_\sigma[\sigma c_\sigma(y)].
\end{align*}
Then for all $s, t\in\mathcal S$, 
\begin{align*}
    \mathbb{E}[g_t(Y) | X=s] & = \mathbb{E}_\sigma[\sigma_t \mathbb{E}[c_\sigma(Y)|X=s]] \\
    & = \mathbb{E}_\sigma[\sigma_t\sigma_s] \\
    & = 1\{s=t\},
\end{align*}
showing that $g$ is unbiased. On the other hand, for any unit vector $u \in \mathbb{R}^S$, 
\begin{align*}
    |\langle u, g(y)\rangle| & = |\mathbb{E}_\sigma[\langle u, \sigma \rangle c_\sigma(y)]| \\
    & \leq \sqrt{\mathbb{E}_\sigma[\langle u,\sigma\rangle^2]} \sqrt{\mathbb{E}_\sigma[c_\sigma^2(y)]} \\
    & \leq 1 + \frac{2}{\gamma}
\end{align*}
where we used Cauchy-Schwarz inequality and the fact that $\mathbb{E}_\sigma[\langle u, \sigma\rangle^2] = \|u\|_2^2 = 1$. Taking supremum over all unit vectors $u$ yields
\begin{align*}
    \sup_{u : \|u\|_2 = 1} |\langle u, g(y)\rangle| = \|g(y)\|_2 \leq 1+\frac{2}{\gamma}
\end{align*}
for all $y \in \mathcal O$. At last, apply \cref{lemma: score-function implies decoder} with $M = \left(1 + \frac{2}{\gamma}\right)^2$ yields a decoder with worst-case success-decoding probability
\begin{align*}
    \Pr(\widehat{X} = s | X=s) \geq \frac{\gamma^2}{(\gamma+2)^2} \geq \frac{\gamma^2}{9} \qquad \text{for all } s\in\mathcal S.
\end{align*}

\section{Belief contraction fails under $\ell_2$-observability} \label{app: counterexample}

In this section, we explain why the belief-contraction approach used in \citep{gol22} does not extend directly to the $\ell_2$-observability setting. Specifically, we construct an example with $\ell_2$-observability constant bounded below by a constant, for which the filters require around $S\log(1/\varepsilon)$ steps to converge to within $\varepsilon$ $\ell_1$-distance (i.e., linear instead of $\mathrm{polylog}S$).

\subsection{Description of the example}
\label{sec: l2 to l1 translation}

Let $G=(\mathcal L,\mathcal R,E)$ be a simple $3$-regular bipartite graph and $|\mathcal L|=|\mathcal R|=n$. We choose $G$ from an infinite family of \emph{bipartite Ramanujan graphs}~\citep{mar13}. The special property of these graphs is that their biadjacency matrix (indicating connections from $\mathcal L$ to $\mathcal R$) $M\in\{0,1\}^{n\times n}$ has every row and column sum equal to $3$ (i.e. $\sigma_1(M)=3$) and satisfies $\sigma_2(M)\leq 2\sqrt{2}$. This constant gap between $\sigma_1$ and $\sigma_2$, denoted by $\nu:=3-2\sqrt{2}>0$ is what we need for the example to work out. We only need such graphs for an unbounded sequence of values of $n$.

The POMDP has $S=2n$ states, one for each vertex, and a single action. Write $\mathcal L=\{L_1,\ldots,L_n\}$ and $\mathcal R=\{R_1,\ldots,R_n\}$. At each transition, the state is resampled uniformly within its current group:
\begin{align}\label{eq: uniform resampling}
    T(L_k\mid L_i)=\frac1n,\qquad T(R_k\mid R_j)=\frac1n
\end{align}
for all $i,j,k\in[n]$, with zero probability of moving between groups. The transition and observation kernels are the same at every stage.

The observation alphabet is $\mathcal O=E\cup\{g_+,g_-\}$, of size $3n+2$. With probability $1/2$, the observation is an edge chosen uniformly from the $d$ edges incident to the current state. Thus, for $e\in E$ and $s\in\mathcal S$,
\begin{align*}
    K(e\mid s)=\frac{1}{6}\mathbf{1}\{s\text{ is an endpoint of }e\}.
\end{align*}
An edge observation restricts the current state to its two endpoints, one in each group.

The remaining probability $1/2$ is assigned to the global symbols $g_+$ and $g_-$. Their probabilities depend only on the group:
\begin{align*}
    K(g_+\mid L_i)&=\frac14+\frac{1}{8\sqrt n}, & K(g_+\mid R_j)&=\frac14-\frac{1}{8\sqrt n},\\
    K(g_-\mid L_i)&=\frac14-\frac{1}{8\sqrt n}, & K(g_-\mid R_j)&=\frac14+\frac{1}{8\sqrt n}.
\end{align*}
Under a uniform state within either group, each edge has probability $1/(6n)$, so an edge observation gives no information about the group. The global symbols provide a weak signal about the group, but no information about the vertex within it.

Let $b_L=\mathrm{Unif}(\mathcal L)$ and $b_R=\mathrm{Unif}(\mathcal R)$. We compare filters initialised at
\begin{align*}
    b_0=b_L,\qquad \widehat b_0=\tfrac12 b_L+\tfrac12 b_R.
\end{align*}
The actual process starts from $b_0$, so its state remains in $\mathcal L$. Write $X_h$ and $Y_h$ for the state and observation at stage $h$. Both filters use the same observation history and the same kernels, but different initial distributions:
\begin{align*}
    b_h(s;y_{1:h})&=\Pr_{b_0}(X_h=s\mid Y_{1:h}=y_{1:h}),\\
    \widehat b_h(s;y_{1:h})&=\Pr_{\widehat b_0}(X_h=s\mid Y_{1:h}=y_{1:h}).
\end{align*}

The channel $K$ has $\ell_2$-observability constant at least $\sqrt{3-2\sqrt2}/6$, independently of $n$; see~\cref{app: proof of l2 observability}. We next show that the two filters nevertheless converge slowly.

\subsection{Slow filter forgetting under $\ell_2$-observability} \label{sec: slow filter forgetting}

Slow forgetting in this example is a hypothesis-testing problem: the second filter must learn whether the state is in $\mathcal L$ or $\mathcal R$. Edge observations do not distinguish the groups. Global symbols occur with probability $1/2$ under either group; conditional on seeing one, the probability of $g_+$ is $1/2+\eta$ under $\mathcal L$ and $1/2-\eta$ under $\mathcal R$, where $\eta:=1/(4\sqrt n)$. Thus the informative observations are coin tosses with opposite biases of order $1/\sqrt n$. We show that the expected filter distance is within constant factors of the optimal error in testing these two hypotheses.

\begin{theorem}\label{thm: no fast filter forgetting}
For the family of POMDPs described above and every $0<\varepsilon\leq1/4$, the number of observations required to reduce $\mathbb E_{b_0}\|b_h-\widehat b_h\|_1$ to at most $\varepsilon$ is $\Theta(S\log(1/\varepsilon))$, provided the horizon is sufficiently large. In particular, constant $\ell_2$-observability does not guarantee a polylogarithmic filter-forgetting time.
\end{theorem}

\begin{proof}
Let $B\in\{L,R\}$ denote the group, which never changes. Conditional on $B$, the states are independent and uniform within that group, so the observations are also independent. Write $P=Kb_L$ and $Q=Kb_R$ for the observation distributions under $L$ and $R$, and $P_h=P^{\otimes h}$, $Q_h=Q^{\otimes h}$ for the distributions of an observation history $z=y_{1:h}$.

The second filter starts with equal probabilities on the two groups. After observing $z$, its posterior probability of $L$ is
\begin{align*}
    \theta_h(z)=\frac{P_h(z)}{P_h(z)+Q_h(z)}.
\end{align*}
Before each observation, both filters are uniform within $\mathcal L$ conditional on being in that group, because the transition resamples the state uniformly. Applying the same observation update therefore gives the same conditional belief within $\mathcal L$. If $\beta_h^L$ and $\beta_h^R$ denote the conditional beliefs within the two groups, then
\begin{align*}
    b_h=\beta_h^L,\qquad
    \widehat b_h=\theta_h\beta_h^L+(1-\theta_h)\beta_h^R.
\end{align*}
Their disjoint supports imply $\|b_h-\widehat b_h\|_1=2(1-\theta_h)$. Since the true group is $L$, averaging over histories gives
\begin{align}\label{eq: expected stage-h filter distance}
    D_h:=\mathbb E_{b_0}\|b_h-\widehat b_h\|_1
    =2\sum_z\frac{P_h(z)Q_h(z)}{P_h(z)+Q_h(z)}.
\end{align}

Let $e_h$ be the minimum probability of error in testing $P_h$ against $Q_h$ with equal prior probabilities. The optimal test chooses the more likely hypothesis, so
\begin{align*}
    e_h=\frac12\sum_z\min\{P_h(z),Q_h(z)\}.
\end{align*}
For $a,b>0$, $\min\{a,b\}\leq2ab/(a+b)\leq2\min\{a,b\}$. Applying this to each summand in~\eqref{eq: expected stage-h filter distance} yields
\begin{align}\label{eq: bounds for expected filter distance}
    2e_h\leq D_h\leq4e_h.
\end{align}
Thus it remains to bound the error of the testing problem.

For equal-prior testing, the standard sample-complexity bound is $\Theta(\log(1/\delta)/\mathrm H^2(P,Q))$ for error $\delta\leq1/8$ and $\mathrm H^2(P,Q)\leq1/8$; see~\citet[Proposition~4.1]{pensia2025testing}. Here $\mathrm H^2(P,Q):=\frac12\sum_y(\sqrt{P(y)}-\sqrt{Q(y)})^2$, and our observation distributions satisfy
\begin{align*}
    \mathrm H^2(P,Q)=\frac12\left(1-\sqrt{1-4\eta^2}\right)=\Theta(1/n).
\end{align*}
By~\eqref{eq: bounds for expected filter distance}, testing error at most $\varepsilon/2$ is necessary, and error at most $\varepsilon/4$ is sufficient, for $D_h\leq\varepsilon$. Both require $\Theta(n\log(1/\varepsilon))=\Theta(S\log(1/\varepsilon))$ observations, proving the claim.
\end{proof}

\subsection{Proof of \texorpdfstring{$\ell_2$}{l2}-Observability in \texorpdfstring{\cref{sec: l2 to l1 translation}}{Appendix of Counterexample}} \label{app: proof of l2 observability}
Denote $u = \frac{1}{\sqrt{2n}}(\mathbf{1}_n, - \mathbf{1}_n)$. Every $v \in \mathbf{1}_{2n}^\perp$ has a unique orthogonal decomposition $v = au + w$ with $a = \langle u, v\rangle$, and $w = (x,y)$ for some $x, y \in \mathbb{R}^n$ that satisfy $\mathbf{1}^{\top}_n x = \mathbf{1}^{\top}_ny = 0$. Next, we compute the 2-norm of $Ku$ and $Kw$, respectively.

Since each edge $e \in E$ has exactly one endpoint in each left and right group,
\begin{align*}
    (Ku)(e) = \sum_s K(e|s) u(s) = \frac{1}{2d}u(L_i) + \frac{1}{2d}u(R_j) = \frac{1}{2d\sqrt{2n}}- \frac{1}{2d\sqrt{2n}}=0.
\end{align*}
On the other hand,
\begin{align*}
    (Ku)(g_+) & = \sum_{i=1}^n K(g_+|L_i) u(L_i) + \sum_{j=1}^n K(g_+ |R_j) u(R_j) = \frac{\sqrt{2}}{8}, \\
    (Ku)(g_-) & = \sum_{i=1}^n K(g_-|L_i) u(L_i) + \sum_{j=1}^n K(g_-|R_j) u(R_j) = -\frac{\sqrt{2}}{8}.
\end{align*}
Combining above,
\begin{align*}
    \|Ku\|_2^2 = \left(\frac{\sqrt{2}}{8}\right)^2 + \left( -\frac{\sqrt{2}}{8}\right)^2 = \frac{1}{16}.
\end{align*}

For $w = (x,y)$, note that $\mathbf{1}^\top_n x = \mathbf{1}^\top_n y = 0$, so
\begin{align*}
    (Kw)(g_+) &= \sum_{i=1}^n K(g_+|L_i) w(L_i) + \sum_{j=1}^n K(g_+|R_j) w(R_j) \\
    &= \left(\frac{1}{4} + \frac{1}{8\sqrt{n}}\right) \mathbf{1}^\top x + \left(\frac{1}{4}- \frac{1}{8\sqrt{n}}\right) \mathbf{1}^\top y = 0, \\
    (Kw)(g_-) &= \sum_{i=1}^n K(g_-|L_i) w(L_i) + \sum_{j=1}^n K(g_-|R_j) w(R_j) \\
    &= \left(\frac{1}{4} - \frac{1}{8\sqrt{n}}\right) \mathbf{1}^\top x + \left(\frac{1}{4} + \frac{1}{8\sqrt{n}}\right) \mathbf{1}^\top y = 0.
\end{align*}
Next, using the regularity of the graph and definition of biadjacency matrix, we may write
\begin{align*}
    \sum_{\{i,j\} \in E} (x_i + y_j)^2 & = \sum_{\{i,j\} \in E} x_i^2 + \sum_{\{i,j\} \in E} y_j^2 + 2 \sum_{\{i,j\} \in E} x_i y_j \\
    &= d\|x\|_2^2 + d\|y\|_2^2 + 2x^\top My \\
    & \geq d(\|x\|_2^2 + \|y\|_2^2) - 2 (2\sqrt{2}) \|x\|_2 \|y\|_2 \\
    & \geq d(\|x\|_2^2 + \|y\|_2^2) - 2\sqrt{2}(\|x\|_2^2 + \|y\|_2^2) \\
    & = (3 - 2\sqrt{2})( \|x\|_2^2 + \|y\|_2^2) \\
    & = \nu \|w\|_2^2.
\end{align*}
Therefore,
\begin{align*}
    \|Kw\|_2^2 & = \sum_{e \in E} \left(\sum_s K(e|s) w(s)\right)^2 \\
    & = \sum_{\{i,j\} \in E} \left(\frac{1}{2d}(x_i + y_j)\right)^2 \\
    & \geq \frac{\nu}{4d^2}\|w\|_2^2.
\end{align*}
The vectors $Ku$ and $Kw$ are supported on the global symbols and the edge symbols, respectively, so they are orthogonal. Since $u$ is a unit vector and $u\perp w$, we also have $\|v\|_2^2=a^2+\|w\|_2^2$. Then for any $v \in \mathbf{1}_{2n}^\perp$,
\begin{align*}
    \|Kv\|_2^2 & = a^2 \|Ku\|_2^2 + \|Kw\|_2^2 \\
    & \geq \frac{a^2}{16} + \frac{\nu}{4d^2} \|w\|_2^2 \\
    & \geq \min\left\{\frac{1}{16}, \frac{\nu}{4d^2}\right\}\left(a^2+\|w\|_2^2\right) \\
    & = \frac{3-2\sqrt{2}}{36}\|v\|_2^2.
\end{align*}
Here we used $d=3$ and $\nu=3-2\sqrt{2}$.
Taking square roots on both sides yields $\|Kv\|_2 \geq \frac{\sqrt{3-2\sqrt{2}}}{6}\|v\|_2$ for all $v \in \mathbf{1}^{\perp}_{2n}$.

\section[Hardness of computing the optimal $\ell_1$-observability constant]{Hardness of computing the optimal $\ell_1$-observability constant} \label{app: hardness of verifying l1-observability}

Given an observation kernel $K \in \mathbb{R}^{O \times S}$, define its optimal $\ell_1$-observability parameter as 
\begin{align*}
    \gamma(K) := \inf_{v \neq 0: v \in \mathbf{1}_S^\perp} \frac{\|Kv\|_1}{\|v\|_1}.
\end{align*}
In this section, we show that computing the optimal $\ell_1$-observability parameter for some observation kernels can be NP-hard.

\begin{definition}[Uniform sparsest-cut value] \label{def: uniform sparsest-cut value}
    Let $G = (V, E)$ be a graph with $n$ vertices. Define the \textit{uniform sparsest-cut value} of $G$ by
    \begin{align*}
        \phi(G) := \min_{\emptyset \neq A \subsetneq V} \frac{|\partial A|}{|A|(n- |A|)},
    \end{align*}
    where $\partial A$ is the set of edges with exactly one endpoint in $A$.
\end{definition}
In general, computing $\phi(G)$ for an arbitrary graph $G$ is NP-hard~\citep{bon12}. We show that for any graph $G$, the observability parameter of (a slight modification) its incidence matrix is the same as its uniform sparsest-cut value.

\begin{theorem}
    Given a graph $G$ with $ n \geq 2$ vertices and $m \geq 1$ edges, one can construct an observation kernel $K_G$ in polynomial time such that 
    \begin{align*}
        \gamma(K_G) = \frac{n}{2m}\phi(G).
    \end{align*}
    In other words, computing the optimal $\ell_1$-observability constant for $K_G$ is NP-hard.
\end{theorem}

\begin{proof}
    Let $G = (V, E)$ be a graph with $n\geq 2$ vertices and $m \geq 1$ edges. For each edge $e \in E$, choose an arbitrary orientation and let $M \in \mathbb{R}^{m \times n}$ be the signed incidence matrix. Let $J = \mathbf{1}_{m,n} \in \mathbb{R}^{m \times n}$ be the all-ones matrix. Define
    \begin{align*}
        K_G = \frac{1}{2m} \begin{pmatrix}
            J+M \\ J-M
        \end{pmatrix} \in \mathbb{R}^{2m \times n}.
    \end{align*}
    Since entries of $M$ belong to $\{-1, 0, +1\}$, $K_G$ is entrywise non-negative. In each column of the stacked matrix, the contributions from $M$ and $-M$ cancel, while the two copies of $J$ contribute $2m$. Thus each column of $K_G$ sums to 1. It follows that $K_G$ is stochastic and could be an observation kernel for a POMDP.

    For $v \in \mathbf{1}_n^\perp$, $Jv= 0$, so for such $v$,
    \begin{align}
        \|K_Gv\|_1 = \left\|\frac{1}{2m} \begin{pmatrix}
            J+M \\ J-M
        \end{pmatrix} v \right\|_1 = \frac{1}{2m} \left\|\begin{pmatrix}
            Mv \\-Mv
        \end{pmatrix} \right\|_1 = \frac{1}{m}\|Mv\|_1. \label{eq: 1-norm of Kv}
    \end{align}

For any non-trivial cut $\emptyset \neq A \subsetneq V$, let $v_A := \mathbf{1}_A - \frac{|A|}{n} \mathbf{1}_n$. Then
    \begin{align*}
        \|Mv_A\|_1= |\partial A| \qquad \text{and} \qquad \|v_A\|_1 = \left(1-\frac{|A|}{n}\right) \cdot |A| + \frac{|A|}{n} \cdot (n-|A|) = \frac{2|A|(n-|A|)}{n}.
    \end{align*}
    By $\ell_1$-observability, $\frac{\|K_Gv_A\|_1}{\|v_A\|_1} \geq \gamma(K_G)$, which implies
    \begin{align*}
        \gamma(K_G) \leq \frac{\frac{1}{m}|\partial A| }{\frac{2|A|(n-|A|)}{n}} = \frac{n}{2m} \cdot \frac{|\partial A|}{|A|(n-|A|)}.
    \end{align*}
    Since this is true for all $\emptyset \neq A \subsetneq V$, taking minimum yields
    \begin{align*}
        \gamma(K_G) \leq \inf_{\emptyset \neq A \subsetneq V} \frac{n}{2m}\frac{|\partial A|}{|A|(n-|A|)} = \frac{n}{2m}\phi(G).
    \end{align*}

We now show the reverse relation. Let $v \in \mathbf{1}_n^\perp$. By relabeling the vertices in $G$, we may assume that $v_1 \leq v_2 \leq \cdots \leq v_n$. Define $d_k := v_{k+1} - v_k \geq 0$ and $A_k:=\{k+1,\ldots, n\}$ for $k \in [n-1]$. Then the zero-sum condition of $v$ implies that we can write $v$ as a weighted-sum of zero-sum vectors that are derived from indicator vectors:
    \begin{align}
        v = \sum_{k=1}^{n-1} d_k \left(\mathbf{1}_{A_k} - \frac{|A_k|}{n}\mathbf{1}_n\right). \label{eq: rewrite v as sum}
    \end{align}
    Therefore, by triangle inequality and non-negativity of $d_k$, we have
    \begin{align}
        \|v\|_1 & \leq \sum_{k=1}^{n-1} d_k \left\|\mathbf{1}_{A_k} - \frac{|A_k|}{n} \mathbf{1}_n\right\|_1 \nonumber \\
        & = \sum_{k=1}^{n-1} d_k \left(|A_k| \cdot \frac{n-|A_k|}{n} + (n-|A_k|) \cdot \frac{|A_k|}{n}\right) \nonumber \\
        & = \frac{2}{n} \sum_{k=1}^{n-1} d_k |A_k|(n-|A_k|). \label{eq: upper bound for 1-norm of v}
    \end{align}
    On the other hand,
    \begin{align*}
        \|Mv\|_1 & = \sum_{\{i,j\} \in E} |v_i - v_j| \\
        & = \sum_{\{i,j\} \in E} \sum_{k=1}^{n-1} d_k \mathbf{1}\{\{i,j\} \in \partial A_k\} \\
        & = \sum_{k=1}^{n-1} d_k \sum_{\{i,j\}\in E} \mathbf{1}\{\{i,j\}\in\partial A_k\} \\
        & = \sum_{k=1}^{n-1} d_k |\partial A_k|.
    \end{align*}
    By definition of $\phi(G)$, $|\partial A_k| \geq \phi(G)|A_k|(n-|A_k|)$ for each $k\in[n-1]$. Hence, by \eqref{eq: 1-norm of Kv},
    \begin{align*}
        \|K_Gv\|_1 & = \frac{1}{m}\|Mv\|_1 = \frac{1}{m}\sum_{k=1}^{n-1} d_k |\partial A_k| \\
        & \geq \frac{\phi(G)}{m}\sum_{k=1}^{n-1} d_k |A_k|(n-|A_k|) \\
        & \geq \frac{n}{2m}\phi(G)\|v\|_1,
    \end{align*}
    where we used \eqref{eq: upper bound for 1-norm of v} in the last inequality. Taking the infimum over nonzero $v\in\mathbf{1}_n^\perp$ yields $\gamma(K_G)\geq \frac{n}{2m}\phi(G)$. Together with the upper bound, this proves the stated equality.
\end{proof}

\newpage

\bibliographystyle{plainnat}
\bibliography{references}
\end{document}